\documentclass[11pt]{article}
\usepackage{graphicx}
\usepackage[]{hyperref}
\usepackage{xcolor}
\usepackage{multirow}
\usepackage{makecell}
\usepackage[sort]{natbib}
\usepackage[ruled,vlined]{algorithm2e}
\usepackage{longtable}

\usepackage[utf8]{inputenc}
\usepackage[OT4]{fontenc}

\begin{document}

\title{Monte Carlo Tree Search: A Review of Recent Modifications and Applications}
\author{Maciej {\'S}wiechowski\footnote{M. {\'S}wiechowski is with QED Software, Warsaw, Poland, email: m.swiechowski@mini.pw.edu.pl}, Konrad Godlewski\footnote{K. Godlewski is with Warsaw University of Technology, Warsaw, Poland, email: jowisz11@gmail.com}, \\ Bartosz Sawicki\footnote{B. Sawicki is with Faculty of Electrical Engineering, Warsaw University of Technology, Warsaw, Poland, email: bartosz.sawicki@pw.edu.pl}, Jacek Ma{\'n}dziuk\footnote{J. Ma{\'n}dziuk is with Faculty of Mathematics and Information Science, Warsaw University of Technology, Warsaw, Poland and AGH University of Science and Technology, Kraków, Poland, email: mandziuk@mini.pw.edu.pl}
}

\date{}

\maketitle
\begin{abstract}
Monte Carlo Tree Search (MCTS) is a powerful approach to designing game-playing bots or solving sequential decision problems. The method relies on intelligent tree search that balances exploration and exploitation. MCTS performs random sampling in the form of simulations and stores statistics of actions to make more educated choices in each subsequent iteration. The method has become a state-of-the-art technique for combinatorial games. However, in more complex games (e.g. those with a high branching factor or real-time ones) as well as in various practical domains (e.g. transportation, scheduling or security) an efficient MCTS application often requires its problem-dependent modification or integration with other techniques. Such domain-specific modifications and hybrid approaches are the main focus of this survey. The last major MCTS survey was published in 2012. Contributions that appeared since its release are of particular interest for this review.
\end{abstract}



\section{Introduction}

Monte Carlo Tree Search (MCTS) is a decision-making algorithm that consists in searching combinatorial spaces represented by trees. 
In such trees, nodes denote states, also referred to as configurations of the problem, whereas edges denote transitions (actions) from one state to another. 

MCTS has been originally proposed in the work by~\cite{uct} and by~\cite{coulom2006efficient}, as an algorithm for making computer players in Go. It was quickly called a major breakthrough~\citep{gelly:go} as it allowed for a leap from 14 \textit{kyu}, which is an average amateur level, to 5 \textit{dan}, which is considered an advanced level but not professional yet. Before MCTS, bots for combinatorial games had been using various modifications of the minimax alpha-beta pruning algorithm~\citep{junghanns1998there} such as MTD(f)~\citep{plaat2014mtd} and hand-crafted heuristics. In contrast to them, the MCTS algorithm is at its core \emph{aheuristic}, which means that no additional knowledge is required other than just rules of a game (or a problem, generally speaking). However, it is possible to take advantage of heuristics and include them in the MCTS approach to make it more efficient and improve its convergence. 

Moreover, in practical applications, the given problem often tends to be difficult for the base variant of the algorithm. The utilitarian definition of ``too difficult'' is that MCTS achieves poor outcomes in the given setting under practical computational constraints. Sometimes, increasing the effective computational budget would help, although in practical applications it may not be possible (e.g., because of strict response times, hardware costs, parallelization scaling). However, often, a linear increase is not enough to tackle difficult problems represented by trees and solving them would require effectively unlimited memory and computational power. 
There can be various reasons for a given problem being hard for MCTS. To name a few - combinatorial complexity, sparse rewards or other kinds of inherent difficulty. Whenever the \textit{vanilla} MCTS algorithm, i.e., implemented in its base unmodified form, fails to deliver the expected performance, it needs to be equipped with some kind of enhancements. In this survey, we will \textbf{focus only on such papers} that introduce at least one modification to the vanilla version of the method. Although  works that describe the use of standard MCTS in new domains have been published, they are not in the scope of this survey.

More recently, MCTS combined with deep reinforcement learning has become the backbone of AlphaGo developed by Google DeepMind and described in the article from~\cite{silver2016mastering}. It has been widely regarded as not only another major breakthrough in Go, but in Artificial Intelligence (AI) in general. It is safe to say that MCTS has been tried in most of combinatorial games and even some real-time video games~\citep{farooq2016starcraft,kim2017opponent}(examples include Poker~\citep{mcts-poker}, Chess~\citep{Silver1140} Settlers of Catan~\citep{mcts-catan}, Othello~\citep{robles2011learning}, Hex~\citep{mcts-hex}, Lines of Action~\citep{winands2010monte}, Arimaa~\citep{arimaa}, Havannah~\citep{havannah} and \emph{General Game Playing} (GGP)~\citep{genesereth2014general,perez2014knowledge}). The latter is particularly interesting because in GGP, computer bots are pitted against each other to play a variety of games without any domain knowledge. 
This is where MCTS is a commonly applied approach.

Driven by successes in games, MCTS has been increasingly often applied in domains outside the game AI such as planning, scheduling, control and combinatorial optimization. We include examples from these categories in this article as well. Formally, MCTS is directly applicable to problems which can be modelled by a Markov Decision Process (MDP)~\citep{lizotte2016multi}, which is a type of discrete-time stochastic control process. Certain modifications of MCTS make it possible to apply it to Partially Observable Markov Decision Processes (POMDP).

\subsection{Methodology}\label{sec:methodology}

We review only such applications of the MCTS algorithm in which the standard version of the algorithm has been deemed ineffective or unsatisfactory, because of the given problem difficulty or unsuitability for the standard version as it is. This is a common scenario, for instance, in real world problems such as Vehicle Routing Problem (VRP) or real-time video games, which tend to be more complex than combinatorial games due to an enormous branching factor, large state spaces, a continuous environment (time and space) and fast-paced action that requires making constant decisions.

It is useful to distinguish between two classes of environments -- single-agent and multi-agent ones.  Let us comment on this using the game domain example. The differences between single-player and multi-player games make various AI techniques suitable for a particular environment. For instance, deterministic single-player games are like puzzles to solve. The success in solving them depends only on the agent and not on external factors such as chance or opponents. Multi-player games require some form of opponent's modelling or cooperation (if they are cooperative). The implementation of MCTS can be optimized appropriately for the given case. 

We focus on multi-agent problems, e.g., multi-player games, with a few exceptions. The exception is that we include such approaches of using the MCTS algorithm that, in our opinion, could be easily applied if the given problem was a multi-agent one. This is due to our subjective evaluation. The idea behind this was to provide readers with relatively well-generalizable modifications and not necessarily tailored for a given single-player game.

Moreover, certain problems may be modelled as single-player or multi-player games. For example, VRP can be modelled as a multi-player cooperative game, in which  each vehicle has its own game tree and the synchronization between them is treated as in a multi-agent system. However, it can also be modelled as a single-player game, in which there is one coordinator player.

This survey covers contributions made after 2012. The reasoning behind 2012 was to start from the year in which the previous comprehensive MCTS survey was published by~\cite{mctsSurvey}. We recommend it  to readers interested in earlier developments related to MCTS and as an additional introduction to the method. Section~\ref{sec:classic_mcts} contains an introduction to the MCTS algorithm as well.

To start with, we focused on \emph{IEEE Transactions on Games} (ToG), which was previously known as \emph{IEEE Transactions on Computational Intelligence and AI in Games} (TCIAG, until 2018).
This is the leading journal concerning scientific, technical, and engineering aspects of games and the main source of high-quality contributions related to MCTS. We have also included the major conference on games - \emph{IEEE Conference on Games} (CoG). It also changed the name and we included proceedings published under the previous name of \emph{IEEE Computational Intelligence and Games} (CIG). Given those two sources, we performed an exhaustive search within each issue of ToG/TCIAG and each proceedings of CoG/CIG and analyzed all articles whether they fit the scope of this survey.

The remaining articles were found through scientific bibliography providers (online libraries): \emph{dblp}, \emph{IEEE Xplore Digital Library} and \emph{Google Scholar}. Using them, we performed search over \textbf{titles} and \textbf{abstracts} by keywords. The following keywords  were chosen: ``Monte Carlo'', ``Monte-Carlo'', ``Tree Search'', ``MCTS'', ``Upper Confidence Bounds'', ``UCT'', ``UCB'', ``Multi-Armed Bandit'', ``Bandit-Based'', ``General Game Playing'' and ``General Video Game Playing''. 
The results were curated by us with the bar set low (i.e., an acceptance rate was high).  Using common sense, we excluded articles which  had not been peer reviewed or were available only from unverified sources. The latter applies to \emph{Google Scholar} which can index such sources too. 

\subsection{Structure of the Survey}\label{sec:structure}

The following survey structure is mainly organized by types of extensions of MCTS algorithm. However, those modifications are usually motivated by challenging applications of the algorithm. These two classifications cross each other multiple times,  which makes unambiguous  navigation  through the survey content troublesome. This section is to present the main structure of the text. The complete overview of references is presented in two Tables, grouped by application and grouped by method. It could be found in the last section Conclusions on pages~\pageref{tbl:applications},~\pageref{tbl:methods1},~\pageref{tbl:methods2}.

The main content of the survey is composed of \textcolor{blue}{seven} sections. 

\vspace{0.5em}
\noindent\textbf{Section~\ref{sec:classic_mcts} - \emph{Classic MCTS}} introduces the MCTS algorithm in its classic (vanilla) variant. The basics are recalled for readers with less experience with the method.

\vspace{0.5em}
\noindent\textbf{Section~\ref{sec:perfect_info} - \emph{Games with Perfect Information}} is devoted to the modifications of MCTS in games with perfect information, which are simpler in the combinatorial aspect. The topic is divided into three subsections:
\begin{description}
\item[Action Reduction] (sec.~\ref{sec:action_reduction}) - extensions based on elimination of possible actions in decision states. Usually, obviously wrong actions are skipped. This could be achieved by manipulation of exploration term in the tree policy function. 
\item[UCT Alternatives] (sec.~\ref{sec:uct_alternatives}) - Upper Confidence Bounds for Trees (UCT) is the core equation in  the MCTS method. Several modification of UCT are discussed.
\item[Early Termination] (sec.~\ref{sec:early_termination}) - techniques for improving efficiency by setting cut-off depth for random playouts.
\item[Opponent Modelling] (sec.~\ref{sec:Opponent_Modelling_PI}) - chosen aspects of modelling opponent's moves in perfect information games
\end{description}

\vspace{0.5em}
\noindent\textbf{Section~\ref{sec:imperfect_info} - \emph{Games with Imperfect Information}} is dedicated to imperfect information games also referred to as games with hidden information. We distinguish six different types of MCTS extensions related to this game genre.
\begin{description}
\item[Determinization] (sec.~\ref{sec:determinization}) - removing randomness from the game by setting a specific value for each unknown feature.
\item[Information Sets] (sec.~\ref{sec:ISMCTS}) - a method to deal with imperfect information games, where states, which are indistinguishable from player's perspective are grouped in information sets. 
\item[Heavy Playouts] (sec.~\ref{sec:heavy}) - extension of simple, random playouts by adding domain-specific knowledge. Methods from this group proved to be effective; however, one should be aware that utilization of domain knowledge is always at the  expense of increased algorithm's complexity.
\item[Policy Update] (sec.~\ref{sec:policy}) – a general group of methods designed to manipulate policy during the tree building.
\item[Master Combination] (sec.~\ref{sec:combination}) - algorithms taking part in competitions are usually a combination of different MCTS extensions.
\item[Opponent Modelling] (sec.~\ref{sec:opponent_modelling}) - modelling the game progress requires modelling the opponent’s actions. This is actually true for all multi-player games,
but is particularly challenging in imperfect information games. 
\end{description}

\vspace{0.5em}
\noindent\textbf{Section~\ref{sec:ml} - \emph{Combining MCTS with Machine Learning}} summarizes the approaches that combine MCTS with machine learning. 
\begin{description}
\item[MCTS + Neural Networks] (sec.~\ref{sec:mcts_nn}) - AlphaGo approach and other inspired methods.
\item[MCTS + Temporal Difference Learning] (sec.~\ref{sec:mcts_tdl}) - MCTS combined with temporal difference (TD) methods.
\item[Advantages and Challenges of Using ML with MCTS] (sec.~\ref{sec:mlchallenges}) - summary of the section.
\end{description}

\vspace{0.5em}

\noindent\textbf{Section~\ref{sec:mcts_evolutionary} - \emph{MCTS with Evolutionary Methods}} summarizes the approaches that combine MCTS with methods inspired by biological evolution. 
\begin{description}
\item[Evolving Heuristic Functions] (sec.~\ref{sec:evolving_heuristic}) - evaluation functions for MCTS are evolved.
\item[Evolving Policies] (sec.~\ref{sec:evolving_policies}) - the aspects related to performing simulations (roll-outs) undergo the evolution process.
\item[Rolling Horizon Evolutionary Algorithm] (sec.~\ref{sec:rhea}) - a competitor algorithm to MCTS with elements inspired by evolutionary algorithms (EA).
\item[Evolutionary MCTS] (sec.~\ref{sec:emcts}) - combination of MCTS and EA in a single approach.
\end{description}

\vspace{0.5em}
\noindent\textbf{Section~\ref{sec:non-games} - \emph{MCTS Applications Beyond Games}} is focused on applications to ``real-world'' problems from outside the games domain. 
\begin{description}
\item[Planning ] (sec.~\ref{sec:planning}) - optimal planning in logistics and robotics. 
\item[Security] (sec.~\ref{sec:security}) - finding optimal patrolling schedules for security forces in attacker-defender scenarios. 
\item[Chemical synthesis] (sec.~\ref{sec:chemical_synthesis}) - planning in organic chemistry domain.
\item[Scheduling] (sec.~\ref{sec:scheduling}) - real business challenges, such as scheduling combined with risk management.
\item[Vehicle Routing] (sec.~\ref{sec:vehicle_routing}) - solutions to various formulations of the Vehicle Routing Problem.
\end{description}

\vspace{0.5em}
\noindent\textbf{Section~\ref{sec:parallelization} - \emph{Parallelization}} discusses parallel MCTS implementations. Both basic approaches and more advanced methods are discussed.

\vspace{0.5em}
\noindent\textbf{Section~\ref{sec:conclusions} - \emph{Conclusions}} presents an overview of the discussed methods and applications in the form of two summary tables, as well as conclusions and open problems.

\vspace{0.5em}

\noindent\textbf{Note} - The majority of the literature works analysed in this survey apply some kind of parameter optimization or fine-tuning.
In particular, the focus is on the \textit{exploration constant} C, the \textit{maximum search depth}, the \textit{efficient use of an computational budget}, or \textit{parameters related to choosing and executing policies (tree policy and default policy)}. Typically such an optimisation is performed empirically.

Due to a variety of possible parameter-tuning approaches and their direct connections to particular implementations of the MCTS mechanism, we decided to address the topic of parameter optimization alongside introduction and discussion of particular MCTS variants / related papers -- instead of collecting them in a separate dedicated section. We believe that such a presentation is more coherent and allows the reader to take a more comprehensive look at a given method.

\subsection{Definitions of the Considered Game Genres}\label{sec:definitions}

In this section we present selected definitions of the most popular genres of games considered in this paper. An application-oriented taxonomy is presented in Table~\ref{tbl:applications} in section~\ref{sec:tables}. 

\paragraph{Perfect Information Games.} Game is classified as perfect information when every player at any time has a complete knowledge about the history and current state of the game. Chess and Go are prominent representatives of this group.

\paragraph{Imperfect Information Games.} 
In games with imperfect information a game state is not fully available to the players, which means that decisions are taken without the full knowledge of the opponent's position. There can be hidden elements of the game, for example cards in hand or face down cards. Game mechanics can include also randomness e.g. drawing a starting hand of cards at the beginning of gameplay.  Most  modern games are classified as imperfect information.

\paragraph{General Game Playing (GGP).} 
GGP~\citep{genesereth2014general} is a research framework that concerns creating computer agents (AI players) capable of playing a variety of games without any prior knowledge of them. Despite a proposal by~\cite{thielscher2010general} specifying how to include games of chance and incomplete information, the official GGP Competition that is organized each year in this area has only involved finite combinatorial games with perfect information. Apart from these three conditions, the games can be of any kind, e.g. they do not need to be only board ones. Also, they can involve any finite number of players including one, i.e., single-player games. The agents have to figure out how to play from scratch, being given only rules of the game to play. This is possible due to the fact that rules are written in a formal logic-based language, called Game Description Language (GDL) discussed in details by~\cite{swiechowski2016fast}, which is a concise way to represent a state machine of the initial state, terminal states and all possible transitions. Therefore, agents are able to simulate games in order to do reasoning. Since 2007, all GGP winners, such as CadiaPlayer developed by~\cite{finnsson2011cadiaplayer} have been simulation-based. An introduction to GGP and review of the related literature can be found in the survey paper by \cite{swiechowskisurvey2015}.

\paragraph{General Video Game Playing (GVGP).} 
GVGP is also referred to as General Video Game AI (GVGAI)~\citep{perez20162014}. The GVGP Competition, first hosted in 2014 at IEEE Conference on Computational Intelligence and Games, has been inspired by a more mature GGP Competition. However, it is not a successor to GGP but rather a different framework where the time to decide on an action is significantly shorter. The goal remains the same - to create agents capable of playing various games based only on rules given to them as a run-time parameter. The games of choice, in GVGP, are simple 2D real-time video games. The examples are Sokoban, Asteroids, Crossfire or Frogger. Similar to GGP, this framework has its own language for game definitions that features concepts such as sprites, levels and it is generally more suitable for said games. The agents have only 40$ms$ to make a move in the game. Despite such a vast difference in response times, both GGP and GVGP competitions are dominated by simulation-based approaches including ones based on MCTS. The competition is open for entries to various specific tracks such as single-player planning, single-player learning, two-player planning and level generation. 

\paragraph{Arcade Video Games.}
Arcade Video Games is a group of simple computer games that gained popularity in the early 1980s. When the MCTS algorithm had been developed, many researchers validated its applicability for games such as Pac-Man, Mario or Space Invaders. In many cases interesting modifications to standard MCTS  were formulated.

\paragraph{Real Time Strategy games.}
\label{sec:rts_games}
In Real Time Strategy games players combat each other through building units and simulating battles on a map. In general, there are two types of units which the players can deploy: economy-based and military. The former  gather resources from certain places on the map; they can be either renewable or depletable depending on the game. These resources are spent on recruiting military units, which fight in battles controlled by the players. This combination of financial and tactical management results in  RTS games posing a significant challenge to tree search algorithms. The three main issues associated with RTS games are:
\begin{itemize}
	\item huge decision space - Starcraft\footnote{Blizzard Entertainment INC.}, the precursor of the genre,  with the branching factor about $10^{50}$ or higher \cite{otanon2013survey}, whereas Chess about 35 and Go roughly 180. Moreover, the player typically has to manage tens of units simultaneously,
	\item real-time - actions can be performed simultaneously, at any time, as opposed to turn-based games where players must take actions one at a time, in turns. Typically, actions in strategy games are divided into macro-management (e.g., strategic decisions) and micro-management (e.g., units' movement). The latter are taken with a higher frequency.
	\item partial observability - the map is covered with \textit{fog of war}, meaning that some parts are hidden from the players. Handling partial observability in RTS games is presented in \cite{uriarte2017single}. Given the current observation and past knowledge the method estimates the most probable game state. This state is called \textit{believe state} and is then sampled while performing MCTS.
\end{itemize}

\section{Classic MCTS}
\label{sec:classic_mcts}

\subsection{Basic Theory}

Monte Carlo Tree Search is an iterative algorithm that searches the state space and builds statistical evidence about the decisions available in particular states. As mentioned in the Introduction, formally, MCTS is applicable to MDPs. MDP is a process modelled as a tuple $(S, A_S, P_a, R_a)$, where:
\begin{itemize}
\item $S$ - is a set of states that are possible in an environment (state space). A specific state $s_0 \in S$ is distinguished as the initial state.
\item $A_s$ - denotes a set of actions available to perform in state $s$. The subscript $S$ can be omitted if all actions are always available in the given environment.
\item $P_{a}(s,s')$ - is the transition function modelled by a probability that action $a$ performed in state $s$ will lead to state $s'$. In deterministic games, the probability is equal to $1$ if the action in state $s$ leads to $s'$, whereas $0$ if it does not.
\item $R_a(s)$ - is the immediate reward (payoff) for reaching state $s$ by action $a$. In Markov games, where states incorporate all the necessary information (that summarizes the history), the action component can be omitted.
\end{itemize}

This formalism enables to model a simulated environment with sequential decisions to make in order to receive a reward after certain sequence of actions. The next state of the environment can be derived only using the current state and the performed action. We will use the game-related term \textit{action} and a broader one - \textit{decision} - interchangeably. The same applies for the term \textit{game} and \textit{problem}.

In non-trivial problems, the state space (e.g. a game tree) cannot be fully searched. In practical applications, MCTS is allotted some computational budget which can be either specified by the number of iterations or the time available for making a decision. MCTS is an \emph{anytime} algorithm. This property means that it can be stopped at anytime and provide the currently best action (decision) using Equation~\ref{eq:move}:
\begin{equation}
\label{eq:move}
a^{*} = \arg\max_{a \in A(s)} Q(s,a)
\end{equation}
where $A(s)$ is a set of actions available in state $s$, in which decision is to be made and $Q(s,a)$ denotes the empirical average result of playing action $a$ in state $s$.
Naturally, the more the iterations, the more confident the statistics are and the more likely it is that the recommended best action is indeed the optimal one.

\begin{figure}[!htb]
\centering
\includegraphics[width=0.9\linewidth]{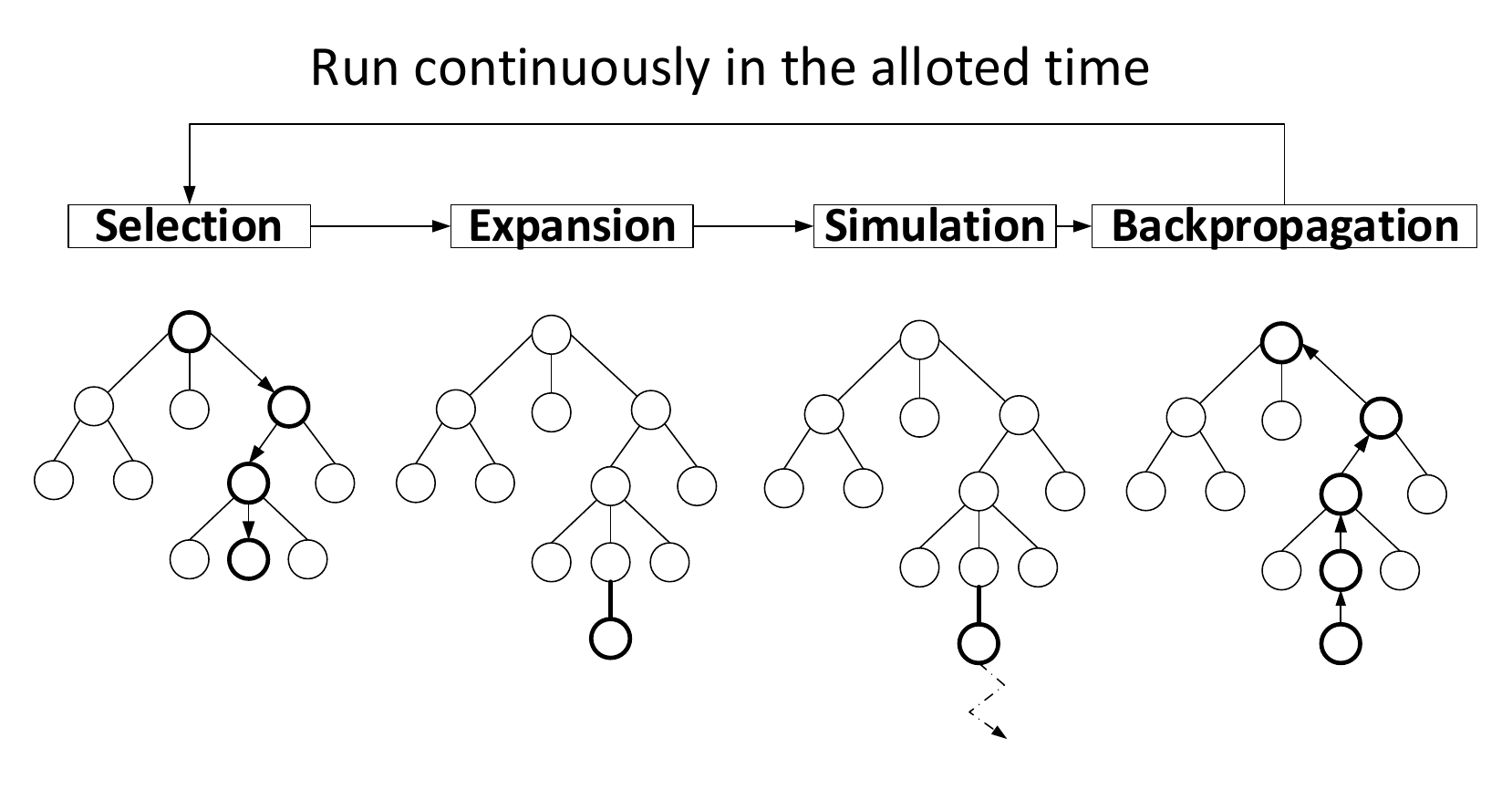}
\caption{Monte Carlo Tree Search phases.} \label{fig:mcts}
\end{figure}

Each iteration consists of four phases as depicted in
Figure~\ref{fig:mcts}:

\begin{enumerate}
	\item \textbf{Selection} - the algorithm searches the portion of the tree that has already been represented in the memory. Selection always starts from the root node and, at each level, selects the next node according to the \emph{selection policy} also referred to as \emph{tree policy}. This phase terminates when a leaf node is visited and the next node is either not represented in the memory yet or a terminal state of the game (problem) has been reached. 
	\item \textbf{Expansion} - unless \emph{selection} reached a terminal state, \emph{expansion} adds at least one new child node to the tree represented in memory. The new node corresponds to a state that is reached by performing the last action in the \emph{selection} phase. When \emph{expansion} reaches the terminal state, which is extremely rare, then the current iteration skips directly to \emph{backpropagation}.
	\item \textbf{Simulation} - performs a complete random simulation of the game/problem, i.e. reaching a terminal state and fetches the payoffs. This is the ``Monte Carlo'' part of the algorithm. 
	\item \textbf{Backpropagation} - propagates the payoffs (scores), for all modelled agents in the game, back to all nodes along the path from the last visited node in the tree (the leaf one) to the root. The statistics are updated.
\end{enumerate}

\subsection{Tree Policy}
\label{sec:tree_policy}

The aim of the selection policy is to maintain a proper balance between the exploration (of not well-tested actions) and exploitation (of the best actions identified so far). The most common algorithm, which has become \emph{de-facto} the enabler of the method is called Upper Confidence Bounds applied for Trees (UCT) introduced by~\cite{uct}. First, it advises to check each action once and then according to the following formula:

\begin{equation}
\label{eq:uct}
a^{*} = \arg\max_{a \in A(s)}\left \{ Q(s,a)+C\sqrt{\frac{ln\left
[N(s)  \right ]}{N(s,a)}} \right \}
\end{equation}

where $A(s)$ is a set of actions available in state $s$, $Q(s,a)$ denotes the average result of playing action $a$ in state $s$ in the simulation performed so far, $N(s)$ is a number of times state $s$ has been visited in previous iterations and $N(s,a)$ - a number of times action $a$ has been sampled in state $s$. Constant $C$ controls the balance between exploration and exploitation. The usually recommended value of $C$ to test first is $\sqrt{2}$, assuming that $Q$ values are normalized to the $[0,1]$ interval, but it is a game-dependent parameter, in general.

\begin{figure}[!htb]
\centering
\includegraphics[width=0.7\linewidth]{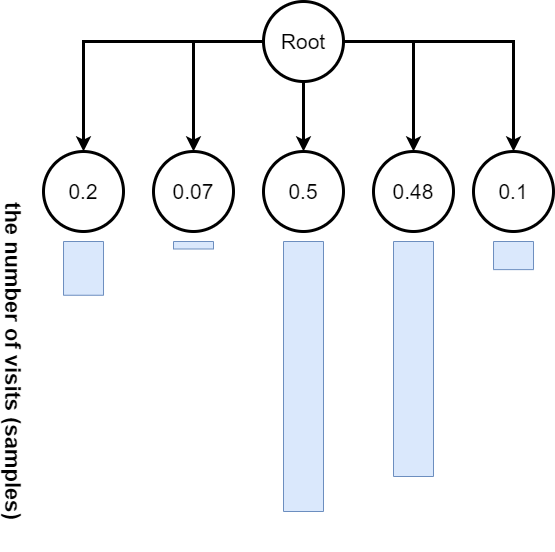}
\caption{The root denotes the starting state. The average scores ($Q(s,a)$) of actions leading to next states are shown inside the circles. The bars denote how many simulations started from a particular action. This is an abstract example - not taken from any particular game.} \label{fig:assymetry}
\end{figure}

Due to UCT formula (c.f.~\ref{eq:uct}) and random sampling, MCTS searches the game tree in an asymmetric fashion in contrast to traditional tree search methods such as minimax. The promising lines of play are searched more thoroughly. Figure~\ref{fig:assymetry} illustrates the type of asymmetrical MCTS search. 

UCT is such a dominant formula for the exploration vs. exploitation trade-off in MCTS, that the approach is often called MCTS/UCT. However, there have been alternatives proposed. One of them is called \emph{UCB1-Tuned}, which is presented in Equation~\ref{eq:ucb1tuned} from the work by~\cite{ucb1tuned}. The only new symbol here compared to Eq.~\ref{eq:uct} is $\sigma_a$ which denotes the variance of the score of action $a$.

\begin{equation}
\label{eq:ucb1tuned}
a^{*} = \arg\max_{a \in A(s)}\left \{ Q(s,a)+C\sqrt{\frac{ln\left
[N(s)  \right ]}{N(s,a)} \min(\frac{1}{4},\sigma_a+\frac{2*ln\left
[N(s)  \right ]}{N(s,a)} )  } \right \}
\end{equation}

The \emph{Exploration-Exploitation with Exponential weights} (EXP3) proposed by~\cite{auer2002finite} is another established tree policy algorithm. In contrast to UCT, it makes no assumption about the reward distribution. For a fixed horizon, EXP3 provides an expected cumulative regret bound against the best single-arm strategy. It is often used in combination with UCT in games with partial observability and/or simultaneous actions.

Thomson sampling is a different than UCT, heuristic approach to the \emph{exploration vs. exploitation} problem. It preceded the MCTS algorithm~\citep{thompson1933likelihood} but has found its use as a tree policy as shown by~\cite{bai2013bayesian}. Thomson sampling is a model that incorporates Bayesian concepts: 
\begin{itemize}
    \item parameterized likelihood function,
    \item prior distribution,
    \item set of past rewards,
    \item posterior distribution.
\end{itemize}  
The model is updated as more information comes in. The expected reward of given action is sampled using the posterior distribution.

Typically, in the \emph{expansion} phase, only one node is added to the tree which is a practical trade-off based on empirical experiments in lots of games. Adding more nodes visited in simulations would slow down the algorithm and increase its memory usage. The further down the road in a simulation a state is visited, the less likely it is for the actual game to reach it. However, if the states (and therefore potential nodes) belonged to the optimal path, it could be beneficial to add them. To make more use of simulations,~\cite{gelly2011monte} proposes a method called \emph{Rapid Value Action Estimation} (RAVE). In RAVE, an additional score is maintained called $Q_{RAVE}$, which is updated for each action performed in the simulation phase and not only for those chosen in the selection steps. The idea is to minimize the effect of the cold start, because especially at the beginning when there are not many samples, the MCTS algorithm behaves chaotically. One of the standard procedures is to linearly weight the RAVE evaluation with the regular average score (c.f. $Q(s,a)$ in Eq.~\ref{eq:uct}) and reduce the impact of RAVE when more simulations have been performed. The new formula is shown in Equation~\ref{eq:rave}:
\begin{equation}
\label{eq:rave}
\sqrt{\frac{k}{3*N(s)+k}} * Q_{RAVE}(s,a) + \left (1-\sqrt{\frac{k}{3*N(s)+k}} \right ) *Q(s,a)
\end{equation}
where $k$ is the so-called RAVE equivalence constant.

There are a few more algorithms that modify or build upon the UCT formula such as \emph{Move-Average Sampling Technique} (MAST) or \emph{Predicate-Average Sampling Technique} (PAST). We recommend papers by~\cite{finnsson2010learning} and~\cite{finnsson2011cadiaplayer} for details.

\subsection{Transposition Tables}

The RAVE algorithm is often used with \emph{Transposition Tables}~\citep{tt}. They have been originally proposed as an enhancement to the alpha-beta algorithm, which reduces size of minmax trees. As shown in paper by~\cite{tt}, within the same computational budget, the enhanced algorithm significantly outperforms the basic one without the transposition tables. The term ``transpositions'' refers to two or more states in the game that can be achieved in different ways. Formally speaking, let $S_i$, $S_j$, $S_k$ be some arbitrary states in the game. If there exists a sequence of actions $(a_i,...,a_k)$, that defines the path from state $S_i$ to $S_k$, and a sequence of actions $(a_j,...,a_k)$ which differs from $(a_i,...,a_k)$ on at least one position, and transforms the state from $S_j$ to $S_k$, then $S_k$ represents transposed states. For simpler management, the MCTS tree is often modeled in such a way, that each unique sequence of actions leads to a state with a unique node in the tree. This may result in a duplication of nodes, even for indistinguishable states. However, one of the benefits of such a duplication, is the fact that, whenever an actual action in the game is performed, the tree can be safely pruned into the sub-tree defined by the state the action lead to. All nodes that are either above the current one or on an alternative branch cannot be visited anymore, so there is no need to store them anymore. The problem is more complicated when transpositions are taken into account, so there is one-to-one mapping between states and nodes. In such a case, the structure is no longer a tree per se, but a directed acyclic graph (DAG). When an action is played in the game, it is non-trivial to decide which nodes can be deallocated and which cannot because they might be visited again. In general, it would require a prediction model that can decide whether a particular state is possible to be encountered again. Storing all nodes, without deallocations, is detrimental not only for performance but mostly for the memory usage, which can go too high very quickly.

\subsection{History Heuristic}

\emph{History Heuristic}, pioneered by the work of~\cite{hh}, is an enhancement to game-tree search that dates back to 1989. It dwells on the assumption that historically good actions are likely to be good again regardless of the state they were performed in. Therefore, additional statistics $Q(a)$ - the average payoff for action $a$ and $N(a)$ - the number of times action $a$ has been chosen, are stored globally. There are a few ways of how these statistics can be used. The most common approach is to use the $Q(a)$ score to bias simulations. In a simulation, instead of uniform random choice, the historically best action is chosen with $\epsilon$ probability, whereas with $1-\epsilon$ probability, the default random choice is performed. This method is called $\epsilon-greedy$. An alternative approach is to use the roulette sampling based on $Q(a)$ or a particular distribution such as Boltzmann Distribution, which is the choice in the article from~\cite{finnsson2010learning}. 
A similar enhancement to History Heuristic is called \emph{Last-Good Reply Policy} (LGRP), which is presented in~\cite{nst}. The idea is to detect countermoves to particular actions and store them globally. For each action, the best reply to it is stored. The authors use the LGRP to rank the unexplored actions in the selection step and the simulation step. In the same article \citep{nst}, the idea is further expanded to the so-called \emph{N-grams Selection Technique} (NST). In NST, the statistics are stored not just for actions but for sequences of actions of length $N$.

\subsection{Section Conclusion}

There is no ``free lunch''~\citep{nofreelunch}, which in the area of MCTS means that most modifications of the base algorithm introduce a bias. Such a bias may increase the performance in a certain problem but decrease in others. There are, however, certain universal enhancements that can be applied such as parallelization. Opening books, discussed e.g. by~\cite{gaudel2010principled}, are another example of a universally strong enhancement, but they are not always available. MCTS with opening books can be considered a \emph{meta-approach}, because the first actions are chosen from the opening book and when, at some point in the game, the available actions are not present in the opening book, the remaining ones are chosen according to the MCTS algorithm. MCTS with relaxed time constraints, e.g. run for days, can even be used to generate the opening books for subsequent runs of the MCTS-based agents~\citep{chaslot2009meta}. Similarly, endgame books can be integrated with MCTS, but this is rarely applied as MCTS is a very strong end-game player and far less efficient at the beginning, where the whole game tree is to be searched.

In this section, we presented the MCTS algorithm as it was originally defined and discussed several standard modifications that were proposed before 2012. In the next sections, we focus exclusively on modifications proposed after 2012.

\section{Games with Perfect Information}
\label{sec:perfect_info}

Almost any modern game features a huge branching factor, which poses a great challenge for search-based algorithms. In recent years, several MCTS modifications intended to tackle this problem such as action abstractions or rapid value estimations have been developed. Most of the modifications can be divided into the three categories:
\begin{itemize}
	\item Action Reduction - limiting the number of available actions,
	\item Early Termination - ending the simulation phase (Fig.~\ref{fig:mcts}) prematurely,
	\item UCT alternatives - other approaches to MCTS selection phase.
\end{itemize}

In the following subsections, those three types of extension will be discussed with several examples. 
Another well-known solution, though recently less common, is to simplify the model of the game. One can construct a prediction model capable of anticipating promising moves in a given situation of the game. \cite{graf2014common} introduce a prediction model for the game of Go. The model features patterns extracted from \textit{common fate graph} - a graph representation of the board \citep{graepel2001learning}. These patterns are updated incrementally during the game both during the tree search and the playouts. This process allows  reduction of the branching factor without affecting the prediction performance.
Model simplification appears to be indispensable in video games. It allowed the use of the MCTS technique for campaign bots in a very complex game - \emph{Total War: Rome II} as discussed in \citep{preuss2020games}.
~\cite{swiechowski2018granular} propose the so-called granular games as a framework for simplification of complex video games in order to apply MCTS to them.














\subsection{Action Reduction}
\label{sec:action_reduction}

In the expansion phase, the MCTS algorithm adds new nodes into the tree for states resulting after performing an action in the game. For many problems, the number of possible actions can be too high. In such a case, the tree is growing sideways, with limited chances for the in-depth inspection of promising branches. The aim of the \textit{action reduction} methods is to limit this effect by eliminating some of the actions.

\cite{sephton2014lordsofwar} apply heuristic move pruning to a strategy card game Lords of War\footnote{Black Box Games Publishing}. The heuristic knowledge is split into two categories: measuring the fitness of a state card-based statistics and evaluating card positions based on heat maps. The former proves more efficient, but the heat maps make use of state extrapolation.

\cite{justesen2014scriptutc} extend the UCT Considering Durations (UCTCD) algorithm for combats in Starcraft~\footnote{Blizzard Entertainment} \citep{churchill2013portfolio}. Unlike UCTCD, where the algorithm searches for unit actions, the proposed method considers sequences of scripts assigned to individual units. The approach is elaborated even further - the experiments prove that the scripts appointed to groups of units are more effective in terms of branching factor reduction.

\begin{figure}[htbp]
	\begin{center}
		\includegraphics[width=0.5\linewidth]{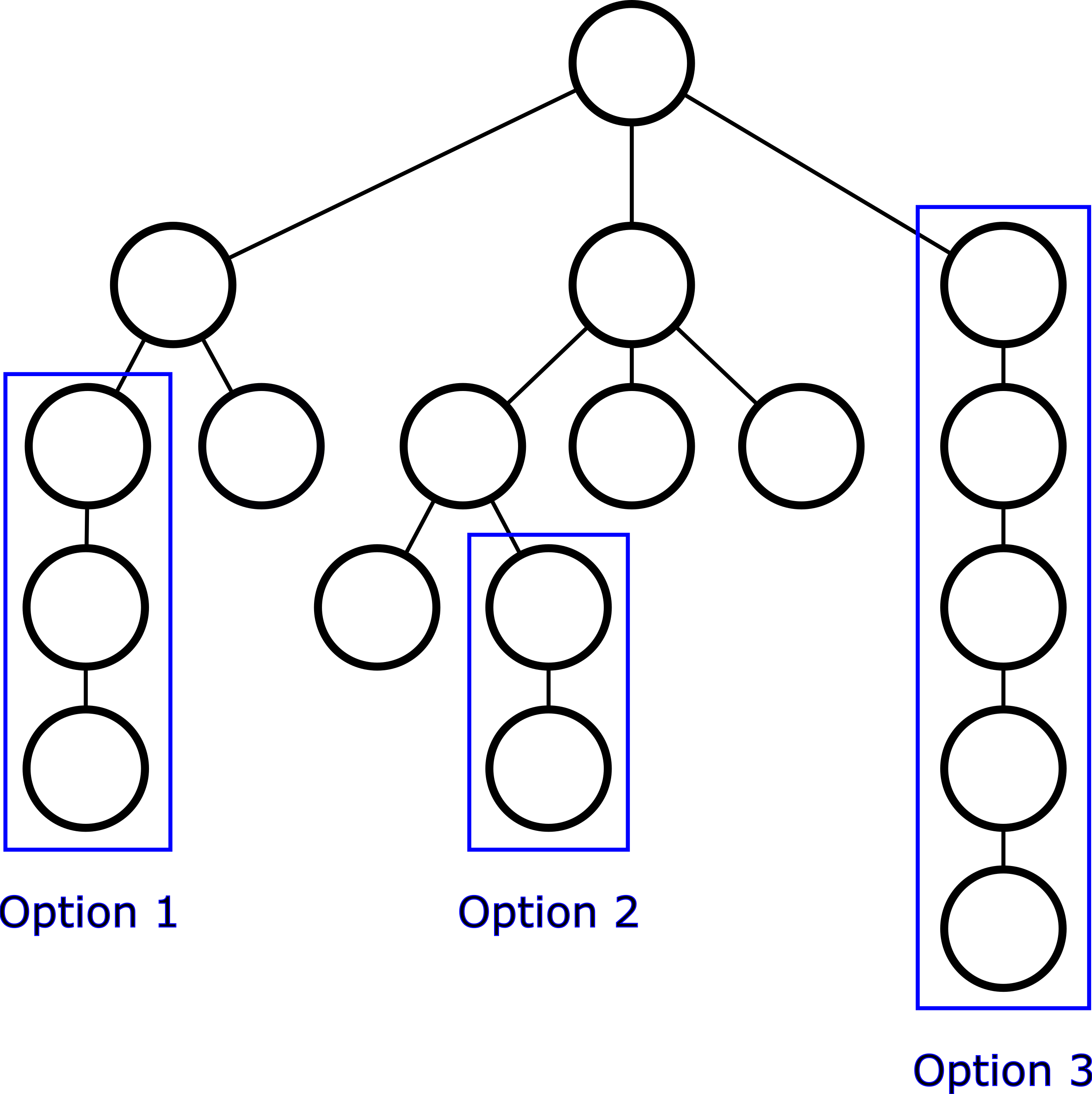} 
		\caption{Representation of options within a tree.} 
		\label{fig:option_new}
	\end{center}
\end{figure}

Similarly to the scripts, Option Monte Carlo Tree Search (O-MCTS) is presented by \cite{de2016monte}. The option is a predefined method of reaching a particular subgoal, e.g. conquering an area in an RTS game. In order to do this, options that contain several consecutive actions within the tree - Fig.~\ref{fig:option_new} were introduced. O-MCTS selects options instead of actions as in vanilla MCTS. Once the selected option is finished (the subgoal has been reached), the expansion phase (Fig.~\ref{fig:mcts}) is conducted. This process allows  a more in-depth search in the same amount of time and diminishes the branching factor.  The paper includes an empirical optimisation of the basic parameters of the algorithm such as discount factor, maximum action time, maximum search depth and UCT constant C. 

Another approach to reducing the branching factor is imposing constraints. Constraints determine situations to be avoided, i.e. actions which result in a defeat, whereas options lead to a specific sub-goal. \cite{subramanian2016efficient} propose a new technique of applying options and constraints to the search policy called Policy-Guided Sparse Sampling (PGSS). PGSS uses constraints for the possibility of pruning a node and options to bias the search towards the desired trajectories. Endowing MCTS with both techniques provides more accurate value estimation. Different action abstraction methods can be found in \citep{gabor2019subgoal,moraes2018action,uriarte2014high}.

Real Time Strategy games are challenging for MCTS due to a huge branching factor. \cite{ontanon2016informed} proposes Informed MCTS (IMCTS) - an algorithm dedicated for games with large decision spaces. IMCTS aggregates information of a game using two novel probability distribution models: Calibrated Naive Bayes Model (CNB) and Action-Type Independence Model (AIM). CNB is a Naive Bayes classifier with a simplified calibration of posterior probability. AIM derives the distribution only for legal actions in the current game state. The models are incorporated into the tree policy by computing the prior probability distributions in the exploration phase. The models are trained on the data from bot duels in an RTS testbed. The experiments show that the models outperform other MCTS approaches in the RTS domain.

\begin{figure}[htbp]
	\begin{center}
		\includegraphics[width=0.6\linewidth]{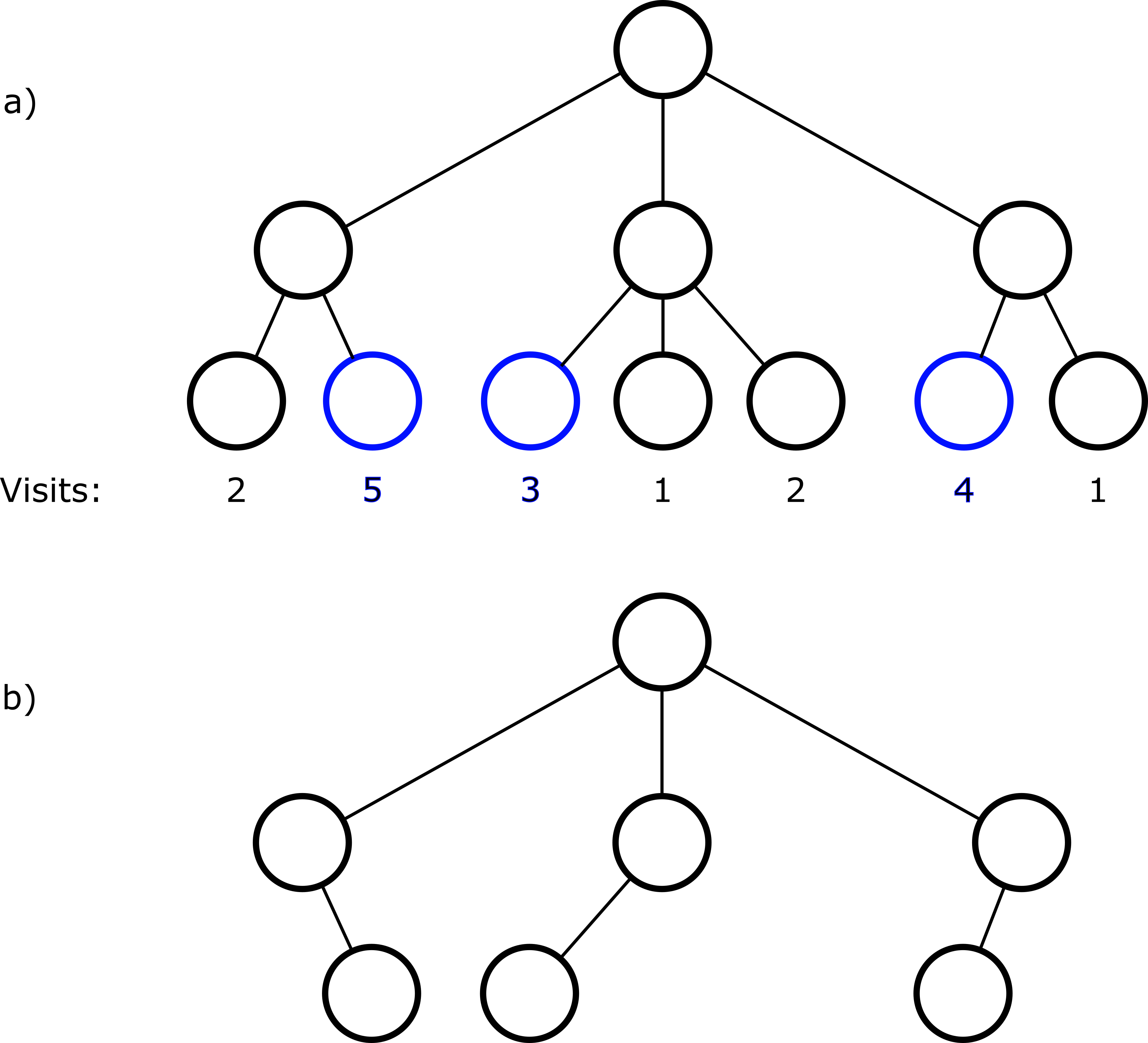} 
		\caption{A sketch of tree pruning at depth \(d=2\) for BMCTS: a) a beam of the 3 most visited nodes (\(W=3\)) is selected, b) the rest is pruned.} 
		\label{fig:beam_new}
	\end{center}
\end{figure}

One of the basic techniques of heuristic action reduction is Beam Search (\cite{lowerre1976harpy}). It determines the most promising states using an estimated heuristic. These states are then considered in a further expansion phase, whereas the other ones get permanently pruned. \cite{baier2012beam} combine Beam Search with MCTS (BMCTS). BMCTS  features two parameters: beam width $W$ and tree depth $d$. They describe how many of the most promising nodes should be selected at an arbitrary depth - Fig. \ref{fig:beam_new}. The nodes outside the beam are pruned. Manual experimentation is used for finding optimal values of parameters: C - exploration coefficient, L - simulation limit, W - beam width.  Different variants of MCTS with a variable depth can be found in \citep{pepels2012enhancements}, \citep{soemers2016enhancements}, \citep{zhou2018hybrid}. 

\cite{ba2019monte} present Variable Simulation Periods (VSP-MCTS). The method extends the decision variables with the simulation time. In VSP-MCTS actions are selected with respect to the possible simulation time before the next selection phase. The available set of action is initially limited to the most promising ones and gradually progressively expanded (unpruned). To achieve this, the authors combine \emph{progressive widening} and \emph{Hierarchical optimistic optimization applied to tree} (HOOT). This approach supports the balance between the frequent action choice and the time dedicated  to simulations.

Progressive Bias \citep{chaslot2008progressive} is another method of heuristic action reduction. Progressive bias extends the UCB function with a heuristic value, which is reduced as the number of node visits increases. \cite{gedda2018monte} propose a progressive win bias:

\begin{equation}
	\label{eq:win_bias}
	P = W \frac{H_i}{T_i (1 - \bar{X_i}) + 1}
\end{equation}
where $H_i$ is the heuristic value,  $\bar{X_i}$ is the average reward for the node, $T_i$ is the number of node visits and $W$ is a positive constant controlling the impact of the bias. Unlike the basic approach, in this formula the heuristic value depends on the number of losses. Other extensions to UCB can be found in \citep{liu2015regulation}, \citep{mandai2016linucb}, \citep{tak2014monte} and \citep{yee2016monte}. \cite{perick2012comparison} compare different UCB selection policies. \cite{ikeda2013efficiency} involve Bradley-Terry Model of static knowledge \citep{coulom2007computing} to bias UCB. A Bias in MCTS can affect the performance in terms of suboptimal moves \citep{imagawa2015enhancements}. 

\cite{demediuk2017monte} introduce a set of \textit{Outcome-Sensitive Action Selection} formulas to create an MCTS agent in a fighting game. The equations evaluate actions with a score function, which in  this case is the difference between the player's and the enemy agent's health. The action with the score closest to zero is selected. 

A large branching factor can be limited by manipulation of the exploration term in the tree policy function (UCT, eq.~\ref{eq:uct}), which is covered in Section~\ref{sec:uct_alternatives}.

\subsection{UCT Alternatives}
\label{sec:uct_alternatives}

In the vanilla MCTS, every state or state-action pair is estimated separately according to the Eq.~\ref{eq:uct} (on page~\pageref{eq:uct}). Some games like the game of Go feature actions which do not strictly depend on the state in which they were played.. To handle this the \textit{all-moves-as-first} (AMAF) heuristic has been developed \citep{brugmann1993monte}. It evaluates each move $a$ by the total reward $\bar{W_a}$ divided by the number of simulations in which $a$ was ever played $\bar{N_a}$.

Rapid Value Estimation, which has already been discussed in section~\ref{sec:tree_policy}, is a generalized idea of AMAF to search trees. It shares action values among subtrees. The value estimation of action $a$ is performed with respect to all the simulations involving $a$ at the subtrees. This process provides more data for each action in comparison to vanilla MCTS, although the additional data may be needless in some cases. In RTS games, any change in positions of nearby units can mislead the action value. The MCTS-RAVE method handles this observation through interpolating the unbiased Monte-Carlo and AMAF value estimate. Thus, the score of state-action pair is as follows~\citep{gelly:go}:

\begin{equation}
	\label{eq:rave_score}
	\bar{Z}(s,a) = (1 - \beta(s,a)) \frac{W(s,a)}{N(s,a)} + \beta(s,a) \frac{\bar{W}(s,a)}{\bar{N}(s,a)},
\end{equation}

where \(\bar{N}(s,a)\) is the AMAF visit count, \(\bar{W}(s,a)\) is the corresponding total reward and \(\beta(s,a)\) is a weighting parameter that decays from 1 to 0 as the visit count \(N(s,a)\) increases. In recent years, several RAVE improvements within the GGP domain have been proposed - \cite{sironi2016comparison}, \cite{cazenave2015generalized}. \cite{sarratt2014converging} and \cite{sironi2019comparing} discuss effects of different parameterization policies on convergence. 

A case analysis of the classical position in the Go game~\citep{browne2012problem} shows that the inclusion of domain knowledge in the form of heavy playouts (see also section~\ref{sec:heavy}) dramatically improves the performance of MCTS, even in the simplest flat version. Heavy playouts reduce the risk of convergence to a wrong move. Exploration constant has been found empirically to give the best convergence. Incorporating domain-independent RAVE technique can actually have a negative effect on performance, if the information taken from playouts is unreliable.

The non-uniformity of tree shape is another serious challenge for the MCTS method. The Maximum Frequency method \citep{imagawa2015enhancements} adjusts a threshold used in recalculation of playout scores into wins or losses. The method is based on maximizing the difference between the expected reward of the optimal move and the remaining ones. The Maximum Frequency method observes the histogram of the results and returns the best threshold based on it. Algorithm parameters are manually adjusted in order to minimize failure rate. The proposed solution has been verified on both random and non-random trees. 

\begin{algorithm}[htbp]
\DontPrintSemicolon
\SetKwProg{Fn}{function}{}{end}
\Fn{MCTS(s, b)}{
\If{s $\notin$ tree}{
	$expand(s)$\;
}
\If{$totalCount(s) > 32$}{
	$b \leftarrow winrate(s)$\;
}
$a \leftarrow select(s)$\;
$s' \leftarrow playMove(s, a)$\;
\eIf{expanded}{
	$(r, g) \leftarrow playout(s')$\;
	$policyGradientUpdate(g, b)$\;
}{
	$r \leftarrow MCTS(s', b)$\;
}
$update(s, a, r)$\;
\Return{r}\;
}
\caption{MCTS with adaptive playouts \citep{graf2016adaptive}}
\label{adaptive_playout_policy}
\end{algorithm}

The playout policy can be changed during a tree search. \cite{graf2016adaptive,graf2015adaptive} show that updating policy parameters (see Alg.~\ref{adaptive_playout_policy}) increases the strength of the algorithm. An adaptive improvement of the policy is achieved by the gradient reinforcement learning technique.

Despite its effectiveness in GGP, there are some domains in which the UCT selection strategy does not work nearly as well. Chess-like games featuring many tactical traps pose a challenging problem for MCTS. One of the tactical issues are optimistic actions - initially promising moves (even leading to a win), but easy to refute in practice \citep{gudmundsson2013sufficiency}. Until MCTS determines the true refutation, the move path leading to an optimistic action is scored favorably. To diminish this positive feedback UCT (see Eq. \ref{eq:uct}) has been extended with \textit{sufficiency threshold} $\alpha$ \citep{gudmundsson2013sufficiency}. It replaces constant $C$ with the following function:

\begin{equation}
	\label{eq:suff_threshold}
	\hat{C} = \left \{\ 
	\begin{array}{ll}
		C & \mbox{when all $Q(s,a) \leq \alpha$},\\
		0 & \mbox{when any $Q(s,a) > \alpha$}.
	\end{array}
	\right.
\end{equation}

When the state-action estimate \(Q(s,a)\) is sufficiently good (the value exceeds \(\alpha\)) then the exploration is discounted; otherwise the standard UCT strategy is applied. This approach allows  rearrangement of the order of moves so that  occasional fluctuations of the estimated value can be taken into account.

Hybrid MCTS-minimax algorithms \citep{baier2014mcts, baier2013winands} have also been proposed for handling domains with a large number of tactical traps. In this context, the basic drawback of vanilla MCTS is the averaging of the sampled outcomes. It results in missing crucial moves and underestimation of significant terminal states. To overcome this, the authors employ shallow-depth minimax search at different MCTS phases (section~\ref{sec:classic_mcts}). Since the minimax search is performed without any evaluation functions, the method does not require domain knowledge and finds its application in any game. Parameters tuning covers exploration constant, minimax depth and visit counter limit.

\subsection{Early Termination} 
\label{sec:early_termination}

MCTS in its classical form evaluates game states through simulating random playouts i.e. complete play-throughs to the terminal states. In dynamic domains, where actions are taken in a short period of time, such outcome sampling appears to be a disadvantage of vanilla MCTS. One of the alternatives is to \textit{early terminate} the playouts - a node is sampled only to an arbitrary depth and then the state is evaluated similarly to the minimax search. One of the key issues is to determine the cut-off depth.  Earlier terminations save the simulation time although they result in evaluation uncertainty. On the other hand, later terminations cause the algorithm to behave more like vanilla MCTS.

\cite{lorentz2016using} finds the optimal  termination point for diverse evaluation functions in the context of three different board games. The results prove that even a weak function can compare favorably to a long random playout. The evaluation functions can also augment the quality of simulations \citep{lanctot2014monte}. The proposed method replaces \(Q(s,a)\) in the classic UCB selection - Eq.~\ref{eq:uct} - with the following formula: 
\begin{equation}
	\label{eq:lanctot_implicit}
	\hat{Q}(s,a) = (1 - \alpha) \frac{r_{s,a}^{\tau}}{n_{s,a}} + \alpha v_{s,a}^{\tau},
\end{equation}
where \(r_{s,a}^{\tau}\) is the cumulative reward of the state-action pair with respect to the player \(\tau\), \(n_{s,a}\) is the visit count, \(v_{s,a}^{\tau}\) is the implicit minimax evaluation for that player and \(alpha\) - weighting parameter. The minimax evaluations are computed implicitly along with standard updates to the tree nodes and maintained separately from the playout outcomes.  Hierarchical elimination tournament with 200 games has been used to tune the parameters of the algorithm. 

The generation of evaluation functions is a challenging research especially in General Game Playing (see section~\ref{sec:definitions}).
\cite{WaledzikMandziukAGI2011,WaledzikMandziukTCIAIG2014} construct a state evaluation function in GGP automatically based solely on the game rules description, and apply this function to guide the UCT search process in their \emph{Guided UCT} method.

Moreover, evaluation functions can be approximated with neural networks \citep{wu2018multilabeled}. \cite{goodman2019re} trains an evaluation network on the data from offline MCTS, which allows  omission of the simulation step during the tree search.

\subsection{Opponent Modelling}
\label{sec:Opponent_Modelling_PI}

Opponent modelling is a challenge that appears in multi-player games. The more information we possess or can infer about the opponent, the better simulation model of their actions we can build. Opponent modelling is a complex topic that is related to games, game theory and psychology. The model of the opponent can be independent of the algorithm an AI agent uses. In this section, we will only scratch a surface and show a few examples of how opponent modelling is incorporated into the MCTS algorithm in the context of games with perfect information.

The paper by~\cite{baier2020guiding} emphasizes the idea of ``focusing on yourself’’ during MCTS simulations. The authors argue that in multi-player games the agent's own actions have often a significantly greater impact on the final score than appropriate opponent prediction. Naturally, there are exceptions to this, such as complex tactical situations. The proposed approach considers abstractions over opponent moves (OMA) that only consider the history (past moves) of the main agent's moves. The abstracted moves share the same estimates that are combined with the UCT formula in various ways. The authors report a better win-rate than vanilla MCTS in six perfect-information, turn-taking, board games with 3+ players.

Fighting games pose a special challenge, since they require a short response time. However, players often repeat the same patterns, e.g., sequences of punches and kicks. An algorithm designed for such a game is described by \cite{kim2017opponent}. An action table is used to describe the opponent's playing patterns. A default action table has three columns depending on the distance to the opponent. Each column corresponds to a particular distance range. After each round the action table is updated based on the observed action frequency.

Opponent modelling is also present in games such as Pacman. In a paper by~\citep{nguyen2012pacman}, the main enhancement was related to the prediction of the opponent's moves, which reduced the number of states analyzed. In the backpropagation phase, the node reward scheme combines the final playout score, but also the simulation time. The playouts are not completely random, the space of the analyzed moves is limited by heuristic rules.














\section{Games with Imperfect Information}
\label{sec:imperfect_info}

The success of MCTS applications to games with perfect information, such as Chess and Go, motivated the researchers to apply it to other games such as card games, real-time strategy (RTS) and other video games. The main feature, which challenges the effective applicability of tree search algorithms is imperfect information. For instance, in RTS games the "fog of war" often appears,  covering the map including the enemy's units.

Imperfect information can increase the branching factor significantly, as every hidden element results in a probability distribution of possible game states. Those states should be included within the tree in the form of additional nodes.

In order to handle imperfect information there have been developed several improvements of the methods such as Perfect Information MCTS (PIMC) (see section \ref{sec:determinization}) and Information Set MCTS (ISMCTS) (see section \ref{sec:ISMCTS}).

Rules of the game and human expert knowledge can be used to increase the reliability of playouts (see Section \ref{sec:heavy} for heavy playouts), or to optimize the tree development process (see Section \ref{sec:policy} for policy update).
The highest quality results can be achieved by combining several methods. Papers of this type usually concern algorithms created for competitions and is described in Section~\ref{sec:combination}.
The last issue related to imperfect information discussed in this section is especially relevant for games and concerns modelling the opponent (see Section~\ref{sec:opponent_modelling}).







\subsection{Determinization} 
\label{sec:determinization}

\begin{figure}[htbp]
	\begin{center}
		\includegraphics[width=0.7\linewidth]{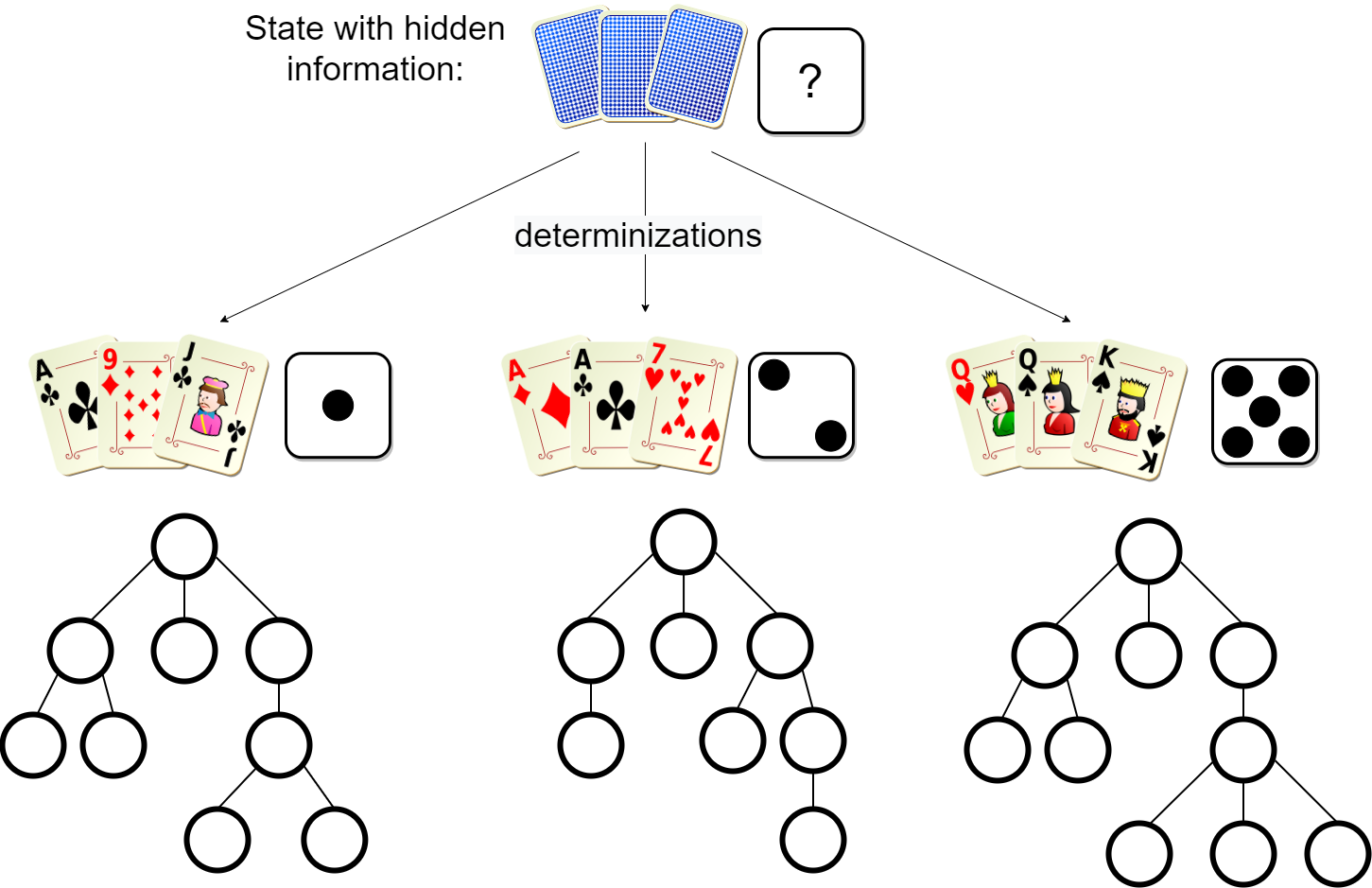} 
		\caption{Determinization method. The dice and the cards in the opponent's hand are sampled in order to form possible subsets. Then each subset rolls out a new subtree.}
		\label{fig:determinizationMaciek}
	\end{center}
\end{figure}

One method of solving the problem of randomness in games is determinization, which  involves sampling from the space of possible states and forming games of perfect information for each such sampling, correspondingly. The sampling requires setting a specific instance (value) for each unknown feature, thus determining it, so it is no longer unknown. For instance, determinization may involve guessing the cards an opponent might have in their hand in a card game (see Fig.~\ref{fig:determinizationMaciek}).

Perfect Information MCTS (PIMC)  assumes that all hidden information is determined and then the game is treated as a perfect information one with respect to a certain assumed state of the world. 
PIMC faces three main challenges \citep{cowling2012infoset}:
\begin{itemize}
	\item strategy fusion - since a given imperfect information state can contain tree nodes with different strategies, a (deterministic) search process may make different decisions in that state depending on particular determinization.
	\item non-locality - optimal payoffs are not recursively defined over subgames,
	\item sharing computational budget across multiple trees.
\end{itemize}

\cite{cowling2012ensemble} apply this method to deal with imperfect information in the game "Magic: The Gathering" (MtG). MtG is a two-person, collectible card game, in which players draw cards  for their starting hand. The cards are not visible to the opponent until they are placed on the table, therefore MtG features partially observable states and hidden information. To approach this game with MCTS, \cite{cowling2012ensemble} propose multiple perfect information trees. This method considerably increases the branching factor, so a few enhancements are employed such as move pruning with domain knowledge and binary representation of the trees. The improvements reduce CPU time per move, rendering the computational budget more manageable.

\subsection{Information Sets} 
\label{sec:ISMCTS}

\begin{figure}[htbp]
	\begin{center}
		\includegraphics[width=0.5\linewidth]{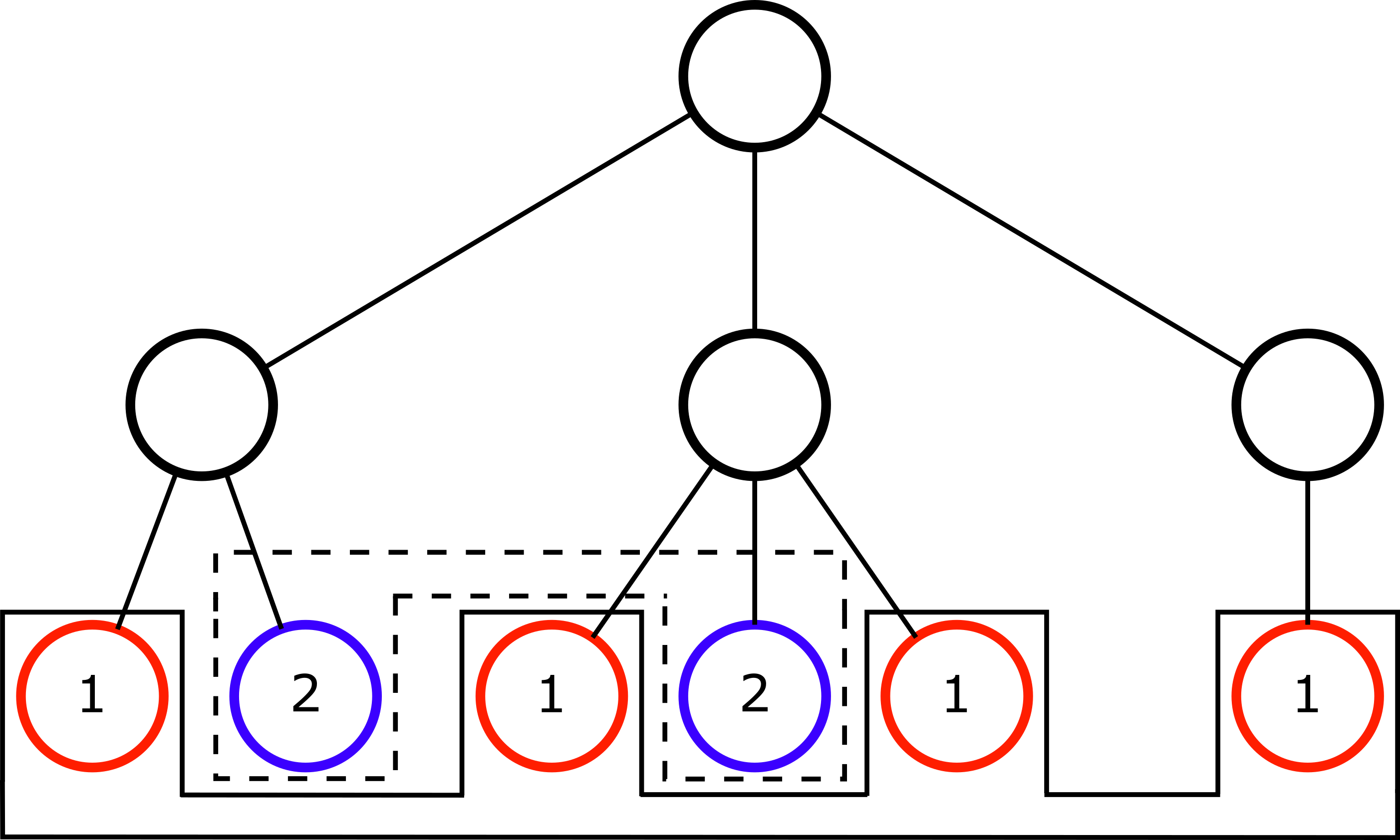} 
		\caption{Two information sets (red and blue). Each set contains nodes indistinguishable from a player's perspective.} 
		\label{fig:infosetFrank}
	\end{center}
\end{figure}

The concept of Information Sets has been introduced by \cite{frank1998search}. It is illustrated in Fig.~\ref{fig:infosetFrank}. The game starts with a chance node, which has three possible moves (circle) for the first player.  The moves of the second player are split into two information sets: red(1) and blue(2). Nodes in each information set are treated as equivalent since the player is only aware of the previous moves, not of the moves below the chance node. 

\cite{cowling2012infoset} propose Information Set MCTS (ISMCTS) to handle the determinization problems: strategy fusion and suboptimal computational budget allocation. The information set is an abstract group of states, which are indistinguishable from a player's point of view \citep{wang2015belief}. Instead of a single state, each node of ISMCTS comprises an information set that allows the player to act upon their own observations. Information sets bring all the statistics about moves in one tree, providing the use of the computational budget in the optimal way. Moreover, applying information sets reduces the effects of strategy fusion, as the search is able to exploit moves that are beneficial in many states of the same information set. For an extension of ISMCTS that incorporates advanced opponent modelling, please refer to Section~\ref{sec:opponent_modelling}.

In games where non-locality is a key factor, \cite{lisy2015online} introduce Online Outcome Sampling (OOS). The algorithm extends Monte Carlo Counterfactual Regret Minimization with two features: building its tree incrementally and targeting samples to more situation relevant parts of the game tree. Performance evaluated on both OOS and ISMCTS shows that only OOS consistently converges close to Nash Equilibrium, and it does so by adapting an offline equilibrium computation to a limited time. This is achieved using the information set about the current position in the game. To tune efficiency of the algorithm sampling technique is used for strategy probabilities parameters.

\cite{furtak2013recursive} introduce Recursive Imperfect Information Monte Carlo (IIMCTS) which is used for playouts with a fixed maximum recursive depth. One of the differences to \cite{cowling2012infoset} is that ISMCTS implicitly leaks game state information by restricting the playouts to be consistent with the player's hidden/private information (e.g. cards in the player's hand in Poker). IIMCTS prevents the leakage by not adapting the players' playout policies across playouts. 

Moreover, the move computation in IIMCTS allows use of the inference mechanism at any node during the lookup, while Information Set MCTS considers nodes only at the root or sets uniform distributions of the hidden state variables. 

Another collectible card game with imperfect information in which MCTS was used is Pokemon \citep{ihara2018pokemon}. In this research, authors compare the effectiveness of determinization versus information sets within MCTS. The results indicate a significant advantage of information sets as ISMCTS reduces the effects of strategy fusion and allocates computing resources more effectively. A similar approach has been presented for Scopone game \citep{di2018traditional}.

\cite{sephton2015experimental} propose the RobustRoulette extension for the selection mechanism in ISMCTS, which averages across many possible states and has a positive effect on the playing strength. 

\cite{uriarte2017single} use a similar method to generate a single believe state - an estimation of the most probable game state. The estimation derives from the current game state, which is not fully observable to the player. This situation is often encountered in Real Time Strategy games (see Section~\ref{sec:definitions}), where enemy units' positions are unknown (covered with ``fog of war''). 

The concept of Information Sets, which is inherent in MCTS application to solving imperfect information problems, has been recently utilized also in domains other than games, e.g. in Security~\citep{KarwowskiMandziukICAISC2015,KarwowskiMandziukEJOR2019,KarwowskiMandziukAAMAS2019,KarwowskiMandziukAAAI2020}. A more detailed discussion is presented in Section~\ref{sec:security}.

\subsection{Heavy Playouts}
\label{sec:heavy}

Heavy playouts is a term defined in the classical paper by \cite{drake2007move}. Authors define two ways of adding domain-specific knowledge into vanilla MCTS: move ordering and heavy playouts. The first one concerns the issue of creating new nodes during the tree's expansion phase, whereas the second one uses a heuristic to bias the ``random'' moves during the rollout.

The implementation of heavy playouts results in more realistic games in the simulation phase. However, it should be remembered that playouts need some degree of randomness to efficiently explore the solution space. Experiments in a card game \emph{Magic: the Gathering} show the importance of the rollout strategy~\citep{cowling2012ensemble}.
A  deterministic expert-rule-based rollout strategy provides an inferior performance to utilizing the reduced-rule-based player that incorporates some randomness within its decisions. This is true even though the expert-rule-based algorithm used as a standalone player is more efficient than the reduced-rule-based algorithm. On the other hand, relying on completely random playouts gives poor performance.

General Game Playing defines a special challenge for the AI agents since they are expected to be able to play games with very different rules (see Section~\ref{sec:definitions}). On the one hand, a fully random playout strategy is universal although, it avoids the use of domain knowledge gained from the past simulations. On the other hand, too much heuristics can lead to omitting a potentially important group of moves. Therefore, one has to find a balance between these  two extremes. \cite{swiechowski2013self} present a method based on switching between six different strategies (from simple random or History Heuristic to more advanced ones). Their solution has been implemented in the MINI-Player agent, which reached the final round in the 2012 GGP Competition. The method was further extended towards handling single-player games more efficiently~\citep{SwiechowskietalTCIAIG2016}.

Genetic programming (GP) has been successfully used to enhance the default MCTS policy in Pac-Man~\citep{alhejali2013pacman}. The GP system operated using seven categories of elements such as IF ELSE statements, numerical operators and comparison operators. Based on these components, a tree policy function was evolved. After sweeping for optimal value of nine different GP and MCTS algorithm parameters, it allowed achievement of a 18\% greater average score than the random default policy. Moreover, the constructed function did not introduce a significant computational overhead. 

Experiment with different 'weights' of heavy playouts are described by \cite{godlewski2021optimisation}. Increasing the amount of expert knowledge added to the playout function generally improves the performance of the AI agent. However, for multistage card games like ``The Lord of the Rings'', each stage might have a different value of optimal playout weight.

\subsection{Policy Update}
\label{sec:policy}

The previous subsection concerned the introduction of knowledge through heavy playouts. In this section, the methods of modification of the tree building policy is  presented. 

The classical RAVE approach (see Section~\ref{sec:tree_policy}) has been extended by \cite{kao2013incentive}. They introduce the RIDE method (Rapid Incentive Difference Evaluation) where the default MCTS policy is updated by using differences (\ref{eq:ride}) between action values for the same state $s$.
\begin{equation}\label{eq:ride}
D^{RIDE}(a,b)=E\{Q(s,a)-Q(s,b)|s \in S \}
\end{equation}
The RAVE and RIDE approaches differ in their sampling strategy. RAVE applies an independent sampling strategy, whereas RIDE applies a pairwise sampling strategy.

The playout policy can be updated by moves, but also based on features of the moves. In the solution presented by \cite{cazenave2016playout}, the tree is developed exactly as UCT does. However, the playouts have weights for each possible move and choose randomly, proportionally to the exponential of the weight. 

Heuristics can be added at the tree development phase or at the playouts phase. Naturally, it is also possible to combine these two approaches~\citep{trutman2015creating}. The authors propose three schemes (playout heuristics, tree heuristics, combined heuristics) in the General Game Playing framework. They have been compared with MAST (Move-Average Sampling Technique)~\citep{finnsson2008simulation}, which is used to bias random playouts, and the RAVE technique for trees. None of the proposed heuristics is universally better than the others and their performance depends on a particular type of game.

The default policy update using heuristics is one of the key enhancements in MCTS experiments for Hearthstone. Since the release in 2014, every year the game has attracted thousands of players, quickly becoming the most popular card game, exceeding 100 million of users worldwide in  2018\footnote{https://videogamesstats.com/hearthstone-facts-statistics/}. 

The game state includes imperfect information - each player's hand is hidden for the opponent. Moreover, the game-play imposes random events and complex decisions. Because of this, state evaluation for this game has always been under research. \cite{santos2017experiments} propose heuristic functions for evaluating subsequent states based on hand-picked features. In addition, they enhance the state search with a database of cards, which contains cards already played by the opponent. The database is used for prediction of cards in the next turn.

State evaluation with a value network \citep{maciek2018improving} is another approach. The authors employ the network to predict the winner in a given state. To transform the state vector as the input of the network, they use skip-gram model instead of hand-picking. A value network from \cite{zhang2017improving} maps a game state to a vector of probabilities for a card to be played in the next turn. Training data for both networks is generated by bot players.

\subsection{Master Combination}
\label{sec:combination}

A special category of research papers are those describing methods developed for game AI competitions~\citep{swiechowski2020game}. Each year, there are many events in which AI players compete against each other. One of the largest ones among academic community members is the IEEE Conference of Games (CoG) (formerly Computational Intelligence and Games - CIG) with over ten categories for popular games such as Angry Birds, Hearthstone and StarCraft; and more general disciplines (General Video Game Playing - GVGP, MicroRTS, Strategy Card Game)~e.g.~\citep{CoG2020}. Open competitions stimulate development of new methods, but particularly good results are achieved by combining multiple methods masterfully. Strong problem-specific enhancements can address different phases of the MCTS algorithm.

One example of such a master combination is the algorithm developed to win the Pac-Man versus Ghost Team competition~\citep{nguyen2012pacman}. The team of ghosts was controlled by MCTS to catch Pac-Man. 
It is  intrinsically interesting that the winner on the Pac-Man side was also an agent using the MCTS algorithm. The set of several improvements was described by \cite{pepels2012enhancements, pepels2014real}.  Most of them refer to heuristic changes in the policy:
\begin{itemize}
\item a tree with variable length edges (taking the distance between decision points into account), 
\item saving the history of good decisions to improve the playout policy, 
\item changes of policy (different modes to eliminate some obviously wrong decisions).
\end{itemize}

\begin{figure}[htbp]
	\begin{center}
		\includegraphics[width=0.6\linewidth]{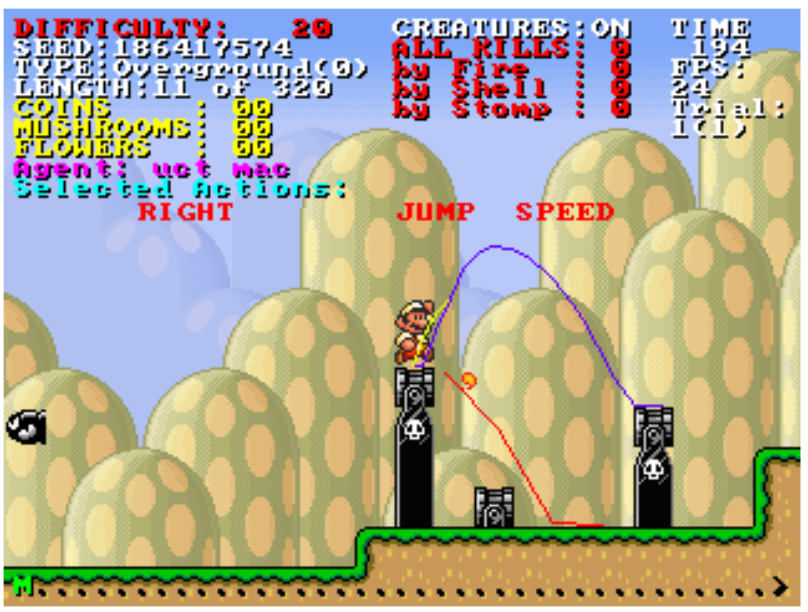} 
		\caption{Comparison of MCTS enhancements for Mario arcade game (Vanilla MCTS - yellow line, Macro Actions - red line, Best combination - blue line). Figure reproduced from \citep{jacobsen2014mario}.}
		\label{fig:jacobsen2014mario}
	\end{center}
\end{figure}

Another popular challenge is the Mario arcade game (see Fig.~\ref{fig:jacobsen2014mario}). \cite{jacobsen2014mario} describe a combination of several improvements to vanilla MCTS that shows excellent results. However, most of them could be classified as an incorporation of domain knowledge into the general algorithm: MixMax (high rewards for good actions), Macro Actions (avoid monsters in a sequence of moves), Partial Expansion (eliminate obvious choices), Hole Detection (extra heuristic to jump over a deadly trap). The paper contains  explanation of optimization process for playout depth and exploration constant.

General Video Game Playing (GVGP) is a class of AI competitions where agents do not know in advance what game will be played (see Section~\ref{sec:definitions}). The vanilla MCTS is an algorithm which does not rely on domain-specific heuristics, so it is naturally a good candidate for such problems. In the first GVG-AI competition at CIG 2014~\citep{CIG2014}, the vanilla MCTS agent surprisingly came in 3rd place, achieving a win-rate of about 32\%.  These impressive results attracted attention to the MCTS method, however, \cite{nelson2016investigating} found out that simply increasing the playout budget is not enough to significantly improve the win-rate. 

\cite{soemers2016enhancements} describe several enhancements of the method, such as N-gram Selection Technique~\citep{tak2012ngrams} - a playout policy update introducing a bias in the steps towards playing sequences of actions that performed well in earlier simulations. The idea of \emph{Loss Avoidance} is particularly interesting, where an agent tries to reduce losses by immediately searching for a better alternative whenever a loss is encountered. A  combination of enhancements allows  an increase in the average win-rate in over 60 different video games from 31.0\% to 48.4\% in comparison with  a vanilla MCTS implementation. 

\cite{frydenberg2015investigating} show that classical MCTS is too cautious for some games in GVGP. Better results are observed after increasing risk-seeking behavior by modification of the exploitation part of UCT. Undesirable oscillations of the controlled character can be eliminated by adding a reversal penalty. Macro-actions and limiting tree expansion are other important improvements. 

Adding a heuristic into the algorithm is a common procedure. \cite{guerrero2017beyond} formulate A conclusion that the best strategy is to change the type of utilized heuristic during the game. They investigate four heuristics: winning maximization, exploration maximization, knowledge discovery and knowledge estimation. It is beneficial to start with enhancing knowledge discovery or knowledge estimation and only then add more winning or exploration heuristics.  These rules could be described by a high level meta-heuristic extension.

An interesting combination of MCTS enhancements is shown by \cite{goodman2019re} describing their efforts to win the Hanabi competition at CIG in 2018. Hanabi is a cooperative game, so communication is an essential factor. The convention used for communication is a particularly interesting sub-problem for AI. When all players are controlled by the same AI agent, hidden information can be naturally shared. In the case of the cooperation between different agents, the communication convention has to be established. For this task, a novel Re-determinizing ISMCTS algorithm has been proposed, supported by adding rules to restrict the action space. The rules of competition give an agent 40ms time for a decision but these were easily achieved using a supervised neural network with one hidden layer.

Nowadays, AI agents usually outperform human players,  which could be perceived as a problem when human enjoyment of the game is necessary. This topic has been studied by \cite{khalifa2016modifying} in terms of GVGP. Several modifications of the UCT formula are introduced, such as an action selection bias towards repeating the current action, making pauses, and limiting rapid switching between actions. In this way, the algorithm behaves in a manner similar to humans, which increases the subjective feeling of satisfaction with the game in human players.

This section can be concluded by stating that the best results are achieved by combining different types of extensions of the MCTS algorithm. However, there is no one-size-fits-all solution. So this should encourage scientists to experiment with various methods.

\subsection{Opponent Modelling}
\label{sec:opponent_modelling}

Problem of opponent modelling is also relevant for games with imperfect information. This section presents a few examples, which incorporate analysis of the opponent into the MCTS algorithm.

\cite{goodman2020does} analyse how accurate the opponent model must be in a simple RTS game called \emph{Ground War}. Two types of opponents are considered - one that uses the same policy as the MCTS-based agent and a parameterized heuristic model. The authors not only have shown that having an accurate opponent model is beneficial to the performance of the MCTS-based agents but also compared its performance against another agent based on Rolling Horizon Evolutionary Algorithm (RHEA). The findings show that the latter is more sensitive to the quality (accuracy) of the opponent model.

A game in which opponent modelling is crucial in order to succeed is Poker. \cite{cowling2015bluffing} extend ISMCTS with the ability to perform inference or bluffing. Inference is a method of deriving the probability distribution of opponent moves based on the observed data (actions already done). However, the opponent can assume that his moves are inferred, and then bluffing comes into play - they can take misleading actions deliberately. Bluffing as an ISMCTS extension is realized by allowing sampling of states that are possible only from the opponent's perspective, the search agent's point of view is neglected. This technique is called self-determinization. The frequency with which the determinization samples each part of the tree is also recorded, thereby allowing performance of inference.

Another example of opponent modelling is Dynamic Difficulty Adjustment (DDA) \citep{hunicke2005case}. DDA is a family of methods focused on creating a challenging environment for the player. In order to keep the player engaged during the whole game, DDA changes the difficulty of encountered opponents according to the player's actual skill level.

A topic which can be related to opponent modelling is making an MCTS-driven player to model a certain behavior. Here, instead of predicting what other players may do in the game, the goal is to make the agent mimic a certain group of players. \cite{baier2018emulating} consider the problem of creating a computer agent that imitates a human play-style in a mobile card game \emph{Spades}. The human-like behavior leads to a more immersive experience for human players, increases retention as well as enables prediction what human players will do, e.g. for testing and QA purposes. The MCTS-based bots are often very strong but rarely feel human-like in their style of playing. The authors train neural networks using approximately $1.2$ million games played by humans. The output of the networks is the probability, for each action, that such an action would be chosen by a human player in a given state. Various network architectures were tested, but all of them were shallow, i.e., not falling into the category of deep learning. Next, various \emph{biasing techniques}, i.e., ways of including the neural networks in the MCTS algorithm were tested. The notable one was based on \emph{knowledge bias}, in which the UCT formula (c.f.~\ref{eq:uct}) was modified as in Eq.~\ref{eq:human_uct}:

\begin{equation}\label{eq:human_uct}
a^{*} = \arg\max_{a \in A(s)}\left \{ Q(s,a)+C\sqrt{\frac{ln\left
[N(s)  \right ]}{N(s,a)}} 
+C_{BT}\sqrt{\frac{K}{N(s)+K}} P(m_i) \right \}
\end{equation}
- where $P(m_i)$ is the output from the neural network trained on human data; $C_{BT}$ is a weight of how the bias blends with the UCT score and $K$ is a parameter controlling the rate at which the bias decreases. Values of $C_{BT}$ and $K$ have been experimentally optimized.

\section{Combining MCTS with Machine Learning}
\label{sec:ml}

\subsection{General Remarks}

Machine Learning (ML) has been on the rise in the computer science research. It has lead to many breakthrough innovations and its popularity and applicability in practical problems is still growing rapidly. The field of ML consists of techniques such as neural networks, decision trees, random forests, boosting algorithms, Bayesian approaches, support vector machines and many others.
Although, Monte Carlo Tree Search is not typically associated with ML, but rather with the classical AI and search problems, it can be considered a machine learning approach, and a type of reinforcement learning, specifically. However, MCTS is often combined with other machine learning models to great success. 
Machine learning, as a general technique, can be used in various ways and contexts. A good example is creating an evaluation function through machine learning. We have already referred to such an approach in Section~\ref{sec:early_termination} and works such as authored by \cite{wu2018multilabeled} and \cite{goodman2019re}. In this section, this topic will be investigated in more details. 
In Section~\ref{sec:imperfect_info}, we have also already shown that convolutional neural networks can be useful tools for map analysis in RTS games~\citep{barriga2017combining}.

\subsection{The AlphaGo Approach - Policy and Value}

The ancient game of \emph{Go} has been considered the next \emph{Grand Challenge} of AI in games since \emph{Chess} and the famous match between Garry Kasparov and Deep Blue, which was analysed by~\cite{campbell2002deep}. Following ~\cite{gelly:go}, \emph{Go} is considerably more difficult for computer programs to play than Chess. Until 2006, computer programs were not stronger than average amateur human player. In 2006-2007, MCTS was introduced and became a quantum leap in terms of bots competence levels. MCTS has become the first major breakthrough in computer \emph{Go}, whereas MCTS combined with machine learning techniques have became the second major breakthrough. In 2015, \emph{AlphaGo} program created by~\cite{silver2016mastering} under the auspices of Google, has become first to beat a professional human player, Fan Hui, on a regular board without a handicap. Subsequently, in 2016, \emph{AlphaGo} won 4-1 in a match versus Lee Sedol~\citep{silver2017mastering}, one of the strongest \emph{Go} players of all time. By many scientists, e.g.~\cite{wang2016does}, this result was considered one of the most important achievements of AI in the recent years. \emph{AlphaGo} inspired many approaches that combine MCTS and ML models. Because many of them share the same basic structure, we will now outline it. The structure is based on two ML models that represent the so-called \textbf{value} and \textbf{policy} functions, respectively. Such models are often referred to as ``heads'' if they are constructed using neural networks, which is the most common choice to modelling them.
\begin{itemize}
\item \textbf{Value function} - a function $v^{*}(s)$ that approximates the outcome of the game in a given state $s$. A perfect (non-approximate) value function is often used when solving a game, e.g. \emph{Checkers} by~\cite{Schaeffer1518}. This is a similar concept to the state evaluation function. The use of them can be technically the same, however, the term ``state evaluation function'' is usually used when the function is static, whereas value function is self-improving through machine learning.
\item \textbf{Policy function} - the policy, denoted by $p(a|s)$ informs which action - $a$ - should be chosen given state $s$. A policy can be deterministic or stochastic. It is often modelled as a probability distribution of actions in a given state.
\end{itemize}

\begin{figure}[!htb]
\centering
\includegraphics[width=4.6in]{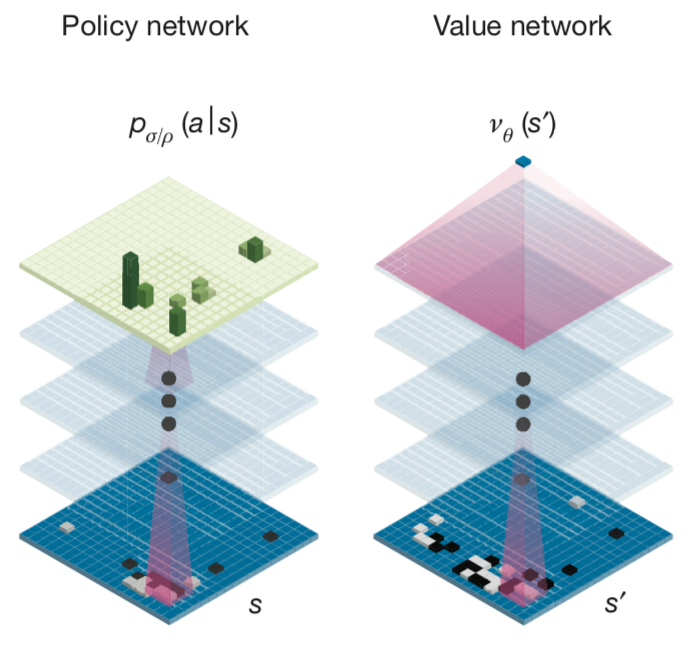}
\caption{Policy and value deep convolutional networks used in AlphaGo. The figure is reproduced from~\citep{silver2016mastering}.}\label{fig:alphago}
\end{figure}

The AlphaGo approach employs deep convolutional networks for modelling both value and policy functions as depicted in Fig.~\ref{fig:alphago}. In contrast to a later iteration of the program called \emph{AlphaZero}, AlphaGo's policy function is kick-started by supervised learning over a corpus of moves from expert human players. Readers interested in the details of the machine learning pipelines pursued in various versions of \emph{AlphaGo} and \emph{AlphaZero} are advised to check the papers from~\cite{Silver1140,silver2016mastering,silver2017mastering}.  The initial policy is called the supervised learning (SL) policy and contains 13 layers~\citep{silver2016mastering}. Such a policy is optimized by a reinforcement learning (RL) policy through self-play and the policy gradient optimization method. This adjusts the SL policy to correctly approximate the final outcome rather than maximize the predictive accuracy of expert moves. Next, the value network is trained to predict the winner in games played by the RL policy.

The authors propose to use a slightly modified formula for selection in MCTS compared to~\ref{eq:uct}:
\begin{equation}
\label{eq:aguct}
a^{*} = \arg\max_{a \in A(s)}\left \{ Q(s,a)+ \frac{P(s,a)}{1+N(s,a))}\right \}
\end{equation}
- where $P(s,a)$ is the prior probability and the remaining symbols are defined in the same way as in Eq.~\ref{eq:uct}.

When a leaf node is encountered, its evaluation becomes:
\begin{equation}
\label{eq:agleaf}
V(s_L)=(1-\lambda )\upsilon_\Theta(s_L)+\lambda z_L  
\end{equation}
- where $\upsilon_\Theta(s_L)$ is the value coming from the value network, $z_L$ is the result of a rollout performed by the policy network and $\lambda$ is the mixing parameter.

\subsection{MCTS + Neural Networks}
\label{sec:mcts_nn}

In this section, we outline approaches that use both the value and the policy functions, so they can be considered as inspired by the AlphaGo approach.~\cite{chang2018big} propose one such work called the ``Big Win Strategy''. It is essentially the AlphaGo approach applied for 6x6 Othello. The authors propose small incremental improvements in the form of an additional network that estimates how many points a player will win/lose in a game as well as using depth rewards during the MCTS search.  

\subsubsection{AlphaGo Inspired Approaches}\label{sec:AlphaGo}

The AlphaGo style training was also applied to the game of \emph{Hex}. \cite{gao2017move} proposed a player named \emph{MoHex-CNN} that managed to outperform the previous state-of-the-art agent named \emph{MoHex 2.0}~\citep{huang2013mohex}. \emph{MoHex-CNN} introduced a move-prediction component based on a deep convolutional network that serves as a policy network and is used together with MCTS. Next, ~\cite{chao} proposed a novel three-head neural network architecture to further significantly improve \emph{MoHex-CNN} and named the new player \emph{MoHex-3HNN}. The network outputs include policy, state- and action-values.


\cite{takada2019reinforcement} present another approach, called \emph{DeepEzo}, that has been reported to achieve a winning rate of $79.3\%$ against \emph{MoHex 2.0}, then world champion. The authors train value and policy functions using reinforcement learning and self-play, as in the case of AlphaGo. The distinctive feature, however, is that the policy function is trained directly from game results of agents that do not use the policy function. Therefore, the policy function training is not a continuous process of self-improvement. Although \emph{DeepEzo} won against \emph{MoHex 2.0}, it lost against \emph{MoHex-3HNN} in a direct competition during the 2018 Computer Olympiad~\citep{gao2019hex}. \\

\cite{yang2020learning} build upon the AlphaGo approach to work without the prior knowledge of \emph{komi}, which is the handicap that can be dynamically applied before a match. 
AlphaGo as well as most of the programs are trained using a komi fixed beforehand, i.e. equal to $7.5$. In the described work, the authors designed the networks that can gradually learn how much komi there is. In addition, they modified the UCT formula to work with the game score directly and not with the win-rate, which is better for unknown komi.~\cite{wu2018multilabeled} propose another approach to tackle the dynamic komi settings. Here, the authors proposed a novel architecture for the value function called a multi-labelled value network. Not only it supports dynamic komi but also lowers the mean squared error (MSE).\\

\cite{tang2016adp} state in the first sentence of the abstract that their method for creating a \emph{Gomoku} agent was inspired by AlphaGo. The main differences are that (1) here only one neural network is trained that predicts the probability of players winning the game given the board state, (2) the network is only three-layer deep (3) the network is trained using \emph{Adaptive Dynamic Programming} (ADP). \\

\cite{barriga2017combining} in their work aim at creating AI agents for RTS games in the $\mu$RTS framework. You can find more details about the approach in the RTS games section~\ref{sec:rts_games} of this survey. The use of machine learning is very similar to the AlphaGo's approach as both value and policy functions are modelled as CNNs. However, the networks are only used for the strategic search, which outputs a strategic action. Next, a purely adversarial search replaces the strategic action by lower level tactical actions. This can be considered a two-level hierarchical approach.

\begin{table}
\centering
\begin{tabular}{|c|c|}
\hline
\textbf{Value network}    & \textbf{Policy network}        \\
Input: 128x128, 25 planes & Input: 128x128, 26 planes       \\
8 conv layers             & 8 conv layers                  \\
\multicolumn{2}{|c|}{Global averaging over 16x16 planes}    \\
2-way softmax             & 4-way softmax                  \\ 
\hline
\end{tabular}
\end{table}

Hearthstone, which has already been introduced in Section~\ref{sec:policy}, is a two player, competitive card game in which the information about game state is imperfect. That partial observability in practice means two situations - when player draws a card and when takes combat decisions without knowing the opponent's cards in hand and the so-called secrets on the board. To handle the first case,~\cite{zhang2017improving} introduce chance event bucketing. Chance event is a game mechanics with a random outcome e.g. a player rolls a dice or draws a card.  Those tens or hundreds different outcomes produced by a single chance event can be divided into an arbitrary number of buckets. Then at the simulation phase each bucket containing similar outcomes is sampled, that process allows to reduce branching factor in the tree search. Other sampling methods based on chance nodes are described by \cite{choe2019enhancing} and \cite{santos2017experiments}.

\cite{maciek2018improving} use the MCTS algorithm together with supervised learning for the game of Hearthstone. In this work, the authors describe various ways in which ML-based heuristics can be combined with MCTS. Their variant of choice uses a value network for the early cutoff which happens at the end of the opponent's turn, therefore the simulation phase of the MCTS algorithm searches the tree only two turns ahead. However, players can make multiple moves within their turns.
The policy network is used to bias the simulations. The authors use pseudo-roulette selection in two variants. In the first one, the probability of choosing an action in a simulation was computed using Boltzmann distribution over heuristic evaluations from the value network. In the second variant, the move with the highest evaluation was chosen with the probability of $\epsilon$, whereas a random one (the default policy) was chosen otherwise, i.e., with the probability of $1-\epsilon$.

Dots-and-Boxes is a paper-and-pencil game that involves capturing territory. For actions, players draw horizontal or vertical lines that connect two adjacent dots. \cite{zhuang2015improving} present an approach to creating a strong agent for this game. The program is based on the MCTS algorithm and a sophisticated board representation. The MCTS is combined with a traditional, i.e., shallow, neural network that predicts the chance of winning of the players. Although this is effectively a value network, it is used to guide the simulation phase of MCTS. The network is trained using stochastic gradient descent. 

\cite{yang2018learning} propose a neural network approach to RTS games in the \textmu RTS framework. It relies on learning map-independent features that are either global (e.g. resources, game time left) or belong to one of 15 feature planes. Each map is represented by a tensor of size $15 \times width \times height$. Such tensor are passed to convolutional neural networks that are trained to estimate winning probability in the current state. 
The generalization ability of the trained network then allows extension to larger maps than the ones used for training.

\cite{8885307} present work that is another of the many examples of approaches that have been inspired by AlphaGo. Here, a system called Spear has been proposed for dependency-aware task scheduling that considers both task dependencies and task resource demands for various resources. It is based on MCTS in which the simulation phase is replaced by a deep reinforcement learning model that acts in the environment and chooses actions according to its policy rather than randomly. 

\subsubsection{MCTS as a Trainer for ML Models}

\begin{figure}[!htb]
\centering
\includegraphics[width=4.6in]{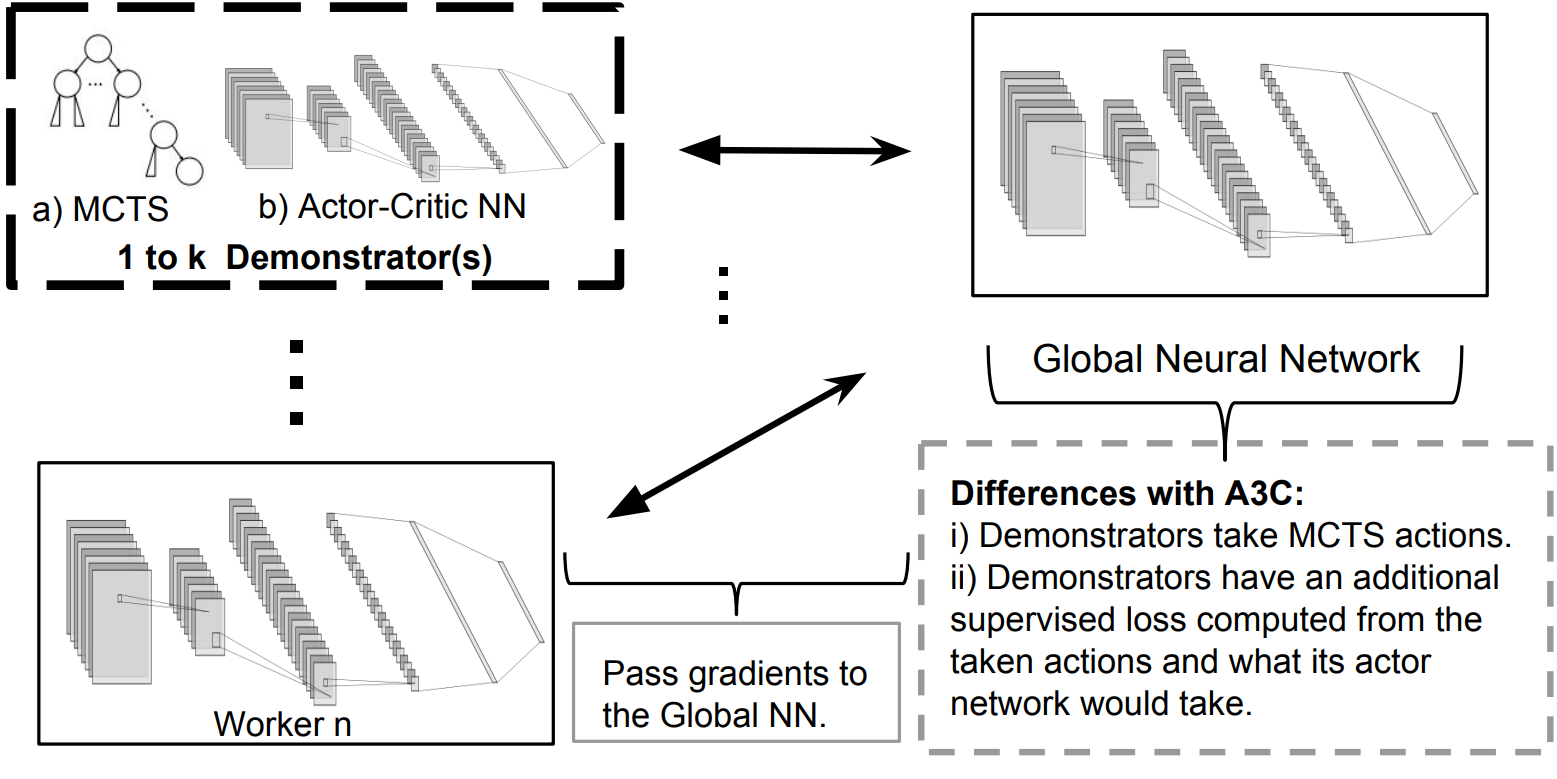}
\caption{Overview of the the Planner Imitation based A3C framework. There are multiple demonstrators making MCTS actions while keeping track of what action the actor network would take. The figure is reproduced from~\citep{kartal2019action_arx}.}\label{fig:pommerman}
\end{figure}

\cite{soemers2019learning} show that is possible to learn a policy in a MPD using the policy gradient method and value estimates directly from the MCTS algorithm. \cite{kartal2019action} propose a method to combine deep reinforcement learning and MCTS, where the latter acts as a demonstrator for the RL component. The authors motivate this approach as ``safer reinforcement learning'' in ``hard-exploration'' domains, where negative rewards may have profound consequences. The domain of choice is a classic video game called \emph{Pommerman}. In each state, the agent may move in one of four directions on the map, stay in place or plant a bomb. The MCTS is run with relatively shallow depth for time-constraints. However it is sufficient to learn, for instance, to avoid being killed by agent's own bomb. Please note that the bombs placed by the agent explode with a delay and this interaction is hard to learn from scratch by pure RL methods.
The approach, depicted in Fig.~\ref{fig:pommerman}, is based on the actor-critic idea. 

A relatively similar idea of using MCTS as a trainer for reinforcement learner was proposed by~\cite{guo2014deep} for simple Atari games. In contrast to the previous approach, the authors use deep convolutional neural networks (CNN) as a model for policy. The input to the CNNs are raw pixel images of what is displayed on the screen. This is essentially the same representation human players observe. The original size of $160$ x $120$ pixels is downgraded to $84$ x $84$ in a preprocessing step. The last four frames, e.g., a tensor of size $84$ x $84$ x $4$ is passed to the network. The output is a fully connected linear layer with a single output for each valid action in the game. The authors compare three different variants of how MCTS can train a CNN: \emph{UCTtoRegression}, \emph{UCTtoClassification} and \emph{UCTtoClassification-Interleaved}.

\cite{pinto2018hierarchical} present another approach in which MCTS and reinforcement learning are combined. Here, MCTS operates in the ``Options Framework''. Options are defined as a tuple $(I, \mu, \beta)$:
\begin{enumerate}
\item $I$ - is the initiation set that contains states from which options are valid;
\item $\mu$ - is a policy that tells what actions to make as part of the current option;
\item $\beta$ - is the termination condition for the option.
\end{enumerate}

The MCTS algorithm searches the combinatorial space of possible options and prepares them for reinforcement learning. 
Using options allowed the authors to reformulate the MPD problem into a semi-markov decision problem (SMDP). In the latter, the Q-learning rule can be defined over options as shown in Eq.~\ref{eq:qsmdp}.
\begin{equation}
\label{eq:qsmdp}
Q(s,o)=(1-\alpha) Q(s,o) + \alpha\left [ r + \gamma \max_{o' \in O}Q(s',o') \right ]
\end{equation}
where $O$ denotes the possible options.

\subsection{MCTS with Temporal Difference Learning}
\label{sec:mcts_tdl}

Temporal Difference (\emph{TD}) learning was one of the first notable applications of the reinforcement learning paradigm in the domain of games~\cite{tesauro1994td}. Here, the game is modeled as time sequences of pairs $(x_t, y_t)$, where $t$ is the time step, $x_t$ denotes a state at this step and $y_t$ is the predicted reward associated with it. The main idea that underpins \emph{TD} learning is that the reward at the given time step can be approximated by a discounted sum of rewards at future time steps:
\begin{equation}
\label{eq:td}
  Y_t = y_{t+1}+ \gamma y_{t+2} + \gamma^2 y_{t+3} + ... 
\end{equation}

The predictions are iteratively updated to be closer to the prediction of future states, which are more accurate. Please note that the last prediction, which is the final game outcome, is the most accurate. The goal is to learn the prediction for the current state or states available from the current states using the training signal from future predictions. The learning process typically runs for many game repeats. This idea is called bootstrapping.

\emph{TD} methods can be applied to any pairs of values, i.e., input and output signals.~\cite{silver2012temporal} use it to learn the weights of an evaluation function components for Go rather than for complete states.

The value function update in \emph{TD} learning is most commonly defined for any state in the game as:
\begin{equation}
V(s_t) \leftarrow  V(s_t) + \alpha(R_{t+1} + \gamma V(s_{t+1}) - V(s_t))    
\end{equation}
where $\alpha$ is the learning factor; $\gamma$ is the discount factor; $R_{t+1}$ is the reward observed in step $(t+1)$ and $V(s_t)$ is the value of state encountered in step $t$ of the game.\\

\cite{vodopivec2014enhancing} incorporate the ideas of \emph{TD} learning to replace the UCT formula (c.f. Eq.~\ref{eq:uct}) and calculate state estimates in nodes in a different fashion. Using \emph{TD} learning inside MCTS affects only the backpropagation phase, in which values are updated, and the selection phase - because the tree policy no longer uses the UCT score. The remaining parts of the MCTS algorithm remain unchanged. The most general formula, which has specific variants in the paper, is based on weighted average of the classic \emph{UCT} estimate and the \emph{TD} estimate:
\begin{equation}
\label{eq:qtduct}
    Q_{TD-UCT} = \omega V + (1-\omega)*Q + C\sqrt{\frac{ln\left
[N(s)  \right ]}{N(s,a)}}
\end{equation}
where $\omega$ is the weight, $V$ is the estimate sampled from the TD value function and the remaining symbols are as defined in Equation~\ref{eq:uct}.

The Equation~\ref{eq:qtduct} is used instead of the UCT formula as the tree policy. The  \emph{TD} value function update ($(V)$) is performed in the back-propagation phase. For this task, the authors specifically chose the \emph{TD}($\lambda$) algorithm, in which a state value $V(s_t)$ is updated based on the immediate reward -- $R_{t+1}$ -- and the value in the following state -- $V(s_{t+1})$:
\begin{equation}
V(s_t) \leftarrow  V(s_t) + \alpha e(s_t) \delta_t   
\end{equation}
where
\begin{equation}
\label{eq:delta_td}
\delta_t = R_{t+1} + \gamma V(s_{t+1}) - V(s_t)
\end{equation}
$\delta_t$ is the temporal difference error (TD error) and $e(s_t)$ denotes the \emph{eligibility trace}. The remaining symbols have already been defined before.

\noindent Three variants of the \emph{TD}($\lambda$) integration have been proposed:
\begin{itemize}
    \item \emph{TD-UCT Single Backup} - here, the authors introduce a simplification that all future values of states have already converged to the final value of the playout. The TD error depends on the distance to the terminal state and the value estimated in a leaf node: $\delta_1 = \gamma^P R_i - V(leaf)$, where $P$ is the number of steps in a playout. 
    \item \emph{TD-UCT Weighted Rewards} - this variant is even more simplified version of \emph{TD-UCT Single Backup} obtained by fixing most of the parameters. In particular, they set $\gamma = 1$ (in Eq.~\ref{eq:delta_td}) and  $\omega = 1$ (in Eq.~\ref{eq:qtduct}) which makes $V$ fully replace the average score estimate $Q$. However, the exploration part of Eq.~\ref{eq:uct} remains unchanged, so this variant combines the idea of TD learning with the UCT exploration part.
    \item \emph{TD-UCT Merged Bootstrapping} - here, there is no longer the assumption that all future state values converge to the final reward of the playout. Instead, this variant includes the regular TD bootstrapping, i.e. the state value is updated based on the value of the next state.
\end{itemize}

In addition, the combinations with the AMAF heuristic and RAVE (c.f. Eq.~\ref{eq:rave}) are considered in the experiments. The various combinations were tested on four games: Hex, Connect-4, Gomoku and Tic-Tac-Toe. One variant, namely \emph{TD-UCT Merged Bootstrapping} with \emph{AMAF}, achieved the best performance in all four games compared to the other variants as well as compared to the plain UCT.\\

The Temporal Difference method can also be used to adapt the policy used in the simulation phase of the MCTS algorithm~\citep{ilhan2017monte}. This is especially useful for games (or other domains) in which uniform random simulation performs poorly. This idea is explored in General Video Game AI games: \emph{Frogs}, \emph{Missile Command}, \emph{Portals}, \emph{Sokoban} and \emph{Zelda}. The Temporal Difference method of choice is Sarsa($\lambda$). 
The authors use a linear value function approximation (as suggested in true online Sarsa($\lambda$)) along with a state representation method. The representation is a vector of numerical features that describe the game state and its weights. Here, the goal is to learn the weights of the state vector and then use a custom rollout procedure that incorporates state evaluation. The outcome of a rollout is calculated as in Eq.~\ref{eq:td}, so the outcome at step $t$ is a discounted sum of future steps. Each consecutive step is performed by a separate rollout. This is a different approach to a vanilla MCTS, in which there is one rollout that fetches the game score, obtained at the terminal state, and propagates the same score to all nodes visited along the path. Here, states on the visited path receive different scores according to the \emph{TD} learning idea. The rollouts are performed according to the $\epsilon-$Greedy selection policy that maximizes the weighted linear combination of features in a state over possible actions that lead to it. The authors achieve better results than vanilla MCTS in 4 out 5 tested games.

\subsection{Advantages and Challenges of Using ML with MCTS}
\label{sec:mlchallenges}

We have reviewed approaches that combine MCTS with machine learning techniques in various ways. Let us conclude this section with a short discussion about the advantages and specific challenges that arise when such a joint approach is pursued. \\

\noindent \textbf{Advantages:}
\begin{itemize}
    \item \textit{New quality results.} The main factor that motivates researchers to introduce ML into MCTS is to obtain better results than the current state-of-the-art. In the game domain, the common goal is to create a next-level player that will win against the strongest programs or human champions. The examples of \emph{AlphaGo}~\citep{silver2016mastering}, \emph{MoHex-3HNN}~\citep{chao} and other show that ML can indeed be an enabler for such spectacular results. In Section~\ref{sec:non-games}, which is devoted to non-game applications, we will also show an approach that combines ML and MCTS and achieves novel results in chemical synthesis~\citep{segler2018planning}.
    \item \textit{Faster convergence.} Another motivation, and ultimately an advantage, that ML brings to MCTS is that it enables MCTS to converge faster. We have seen three groups of approaches in this category. First,  e.g. by~\cite{zhuang2015improving},~\cite{wu2018multilabeled} and~\cite{maciek2018improving}, achieve it by introducing an ML model for the evaluation function in MCTS. Second, which is often combined with the first as in \emph{AlphaGo}'s inspired approaches, consists in using an ML model as the policy for simulations. Lastly, RL algorithms such as \emph{TD}($\lambda$)  can be combined with the UCT formula or used instead in MCTS as shown by~\cite{vodopivec2014enhancing}.
    \item \textit{Approximation capabilities.} Let us discuss the last two advantages on a theoretical level. MCTS is a search-based method. In order to properly evaluate the quality of a state, the state must be visited enough number of times (to gain the confidence about the statistics). The MCTS algorithm has no inherent ability to reason about the states rather than based on statistics stored directly for them. Many ML models, however, such as neural networks or logistic regression are universal approximators. They can provide answers for states that have never been observed during simulations as long as the ML models are well-trained, i.e., achieve good performance on the training data and are neither underfit nor overfit. 
    \item \textit{Generalization.} Continuing on the aspect the approximation - ML models can generalize to some extend the knowledge encoded in MCTS trees. Although the training process usually takes a long time, using the trained models in real-time is by orders of magnitude faster than using MCTS or any search-based method. Moreover, let us imagine a solved end-game position in Chess, in which there is a strategy for a king that involves a traversal around an enemy piece. If we increased the board size from 8x8 to 9x9, then MCTS not only would have to compute everything from scratch but also the complexity of the problem would increase (bigger board). Thanks to generalization and abstraction capabilities, ML models can bring new quality here.
\end{itemize}
\noindent \textbf{Challenges:}
\begin{itemize}
    \item \textit{Data volume and quality.} The biggest challenge of introducing ML into MCTS is data availability. ML models, especially those based on deep learning such as \textit{AlphaZero}~\citep{silver2017mastering} require a lot of data. Moreover, there is a saying ``\textit{garbage in, garbage out}'' so the data need to be of high quality. In game AI, it means that it must be generated by strong AI players. However, if ML is to help MCTS in the development of strong players, then we have a situation that strong players are not yet available. This is a major obstacle for using supervised learning algorithms. To tackle this problem, self-play and reinforcement learning~\citep{kartal2019action} can be applied.
    \item \textit{Training accurate models.} This is a general challenge when creating ML models. In game AI, overfitting may appear, for example, when heuristic-based and non-adaptive AI players are used in training games to generate data. It may lead to a scenario, in which a lot of similar states are present in the training data, whereas a lot of areas are under-represented. A proper variety of positions must be ensured in the training process as well as proper training methodology. Often, experts at game AI, especially in video game development industry, are not experts in machine learning.
    \item \textit{Technology problems.} The most common programming languages used for games are \emph{C}++ and \emph{C}\#. The most common programming language used for machine learning is \emph{Python}, followed by \emph{R}. Therefore, using ML techniques in games usually requires either custom implementations of the ML models or having complex projects that combine multiple programming languages.
\end{itemize}

\section{MCTS with Evolutionary Methods}
\label{sec:mcts_evolutionary}


\subsection{Evolving Heuristic Functions}
\label{sec:evolving_heuristic}

\begin{figure}[!htb]
\centering
\includegraphics[width=3.8in]{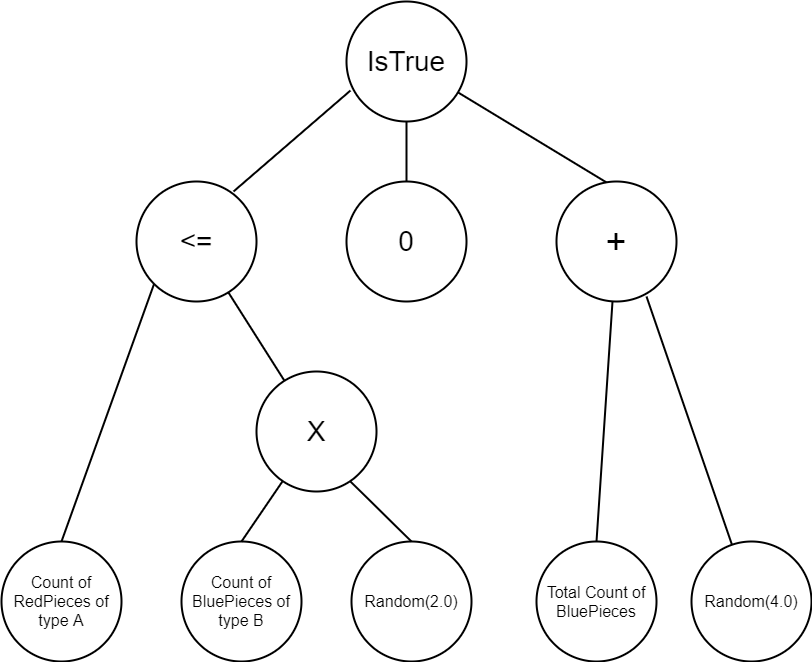}
\caption{An example of a genetic programming tree used to evolve an evaluation function. The figure was inspired by a depiction of the evaluation function for Reversi in work by~\cite{benbassat2013evomcts}.}\label{fig:genprog}
\end{figure}

\cite{benbassat2013evomcts} describe an approach, in which each individual in the population encodes a board-evaluation function that returns a score of how good the state is from the perspective of the active player. The authors use strongly typed genetic programming framework with boolean and floating-point types. The genetic program, i.e., the evaluation function, is composed of terminal nodes and operations such as logic functions, basic arithmetic functions and one conditional statement. The terminal nodes consist of several domain-independent nodes such as \emph{random constant}, \emph{true}, \emph{false}, \emph{one} or \emph{zero} as well as domain-specific nodes such as \emph{EnemyManCount}, \emph{Mobility} or \emph{IsEmptySquare(X,Y)}. The paper considers two domains of choice: \emph{Reversi} and \emph{Dodgem} games. Figure~\ref{fig:genprog} illustrates exemplar function for Reversi.

The authors incorporate the evolved heuristics in the simulation phase of MCTS. Instead of simulating a random move, the agent randomly chooses a parameter called \emph{playoutBranchingFactor} in each step. The upper bound of this value is equal to the number of currently available actions. Then, a set of \emph{playoutBranchingFactor} random moves is selected and the one with the highest heuristic evaluation is chosen to be played in the simulation. This way, there is still stochastic element in simulations, which is beneficial for the exploration of the state-space.

The authors conclude that evolving heuristic evaluation function improves the performance of the MCTS players in both games. However, it is time consuming, therefore it is most suitable in scenarios, when it can be done offline, i.e., before the actual game, which is played by MCTS with an already prepared evaluation function.

\cite{alhejali2013pacman} propose a similar idea to evolving heuristics by means of genetic programming for \emph{Ms Pac-Man} agent. In this research, the new default policy to carry out simulations is evolved directly.

\subsection{Evolving Policies}
\label{sec:evolving_policies}

The approaches presented in this section consist in optimizing the policies of MCTS -- tree policy or default simulation policy -- using evolutionary computation. This is a different case to the previous section, whereby a new evaluation function is constructed. 

One of the first papers on this topic was published by~\cite{lucas2014fast}. The authors introduce a weight vector $w$ that is used to influence both tree policy $T(w)$ and default policy $D(w)$. The weight vectors are stored individuals optimized by a (1+1) Evolution Strategy (ES).
For default policy, a mapping from the state space to a feature space with $N$ features is introduced. Those features are assigned weights that are used to bias actions during a simulation towards states with a greater aggregated sum of weights. To maintain exploration, softmax function is used instead of a greedy selection.

\cite{perez2014knowledge} introduce an approach called \emph{Knowledge-Based Fast Evolutionary MCTS (KB Fast-Evo MCTS)} which extends the work by~\cite{lucas2014fast}. The environment of choice is General Video Game Playing, which as a multi-domain challenge, encourages adaptive algorithms that are not tailored for a particular game. The authors dynamically extract features from a game state, in contrast to a fixed number of features employed in the previous work. The features not only vary between games but also between steps of the same game. Each feature $i$ has a weight $w_{ij}$ assigned for action $a_i$. The weights are evolved dynamically and they are used to bias the MCTS simulations in a similar way as proposed by~\cite{lucas2014fast}. Here, also a soft-max function is used to calculate the probability of choosing an action.

\emph{KB Fast-Evo MCTS} also introduces a dynamic construction of a knowledge base. The knowledge base consists of knowledge items, for which statistics such as occurrences and average scores changes are maintained. The authors introduce the notions of \emph{curiosity} and \emph{experience} and the formula to calculate them for a knowledge base. Next, both the values, i.e. \emph{curiosity} and \emph{experience}, are combined to calculate the \emph{knowledge change} for a knowledge base. Finally, the authors propose a formula to calculate the predicted end-game reward that can be used for the MCTS roll-out. The reward is either equal to the change of the game score or equals to the weighted sum of the \emph{knowledge change} and the \emph{distance change} if no game score change is observed. The \emph{distance change} is a heuristic value introduced for GVGP. This approach was tested using 10 games from the GVGP corpus. The proposed \emph{KB Fast-Evo MCTS} outperformed the vanilla MCTS in all of the games, whereas the previous \emph{Fast-Evo MCTS} in 7 out of 10 games.

\begin{figure}[!htb]
\centering
\includegraphics[width=4.6in]{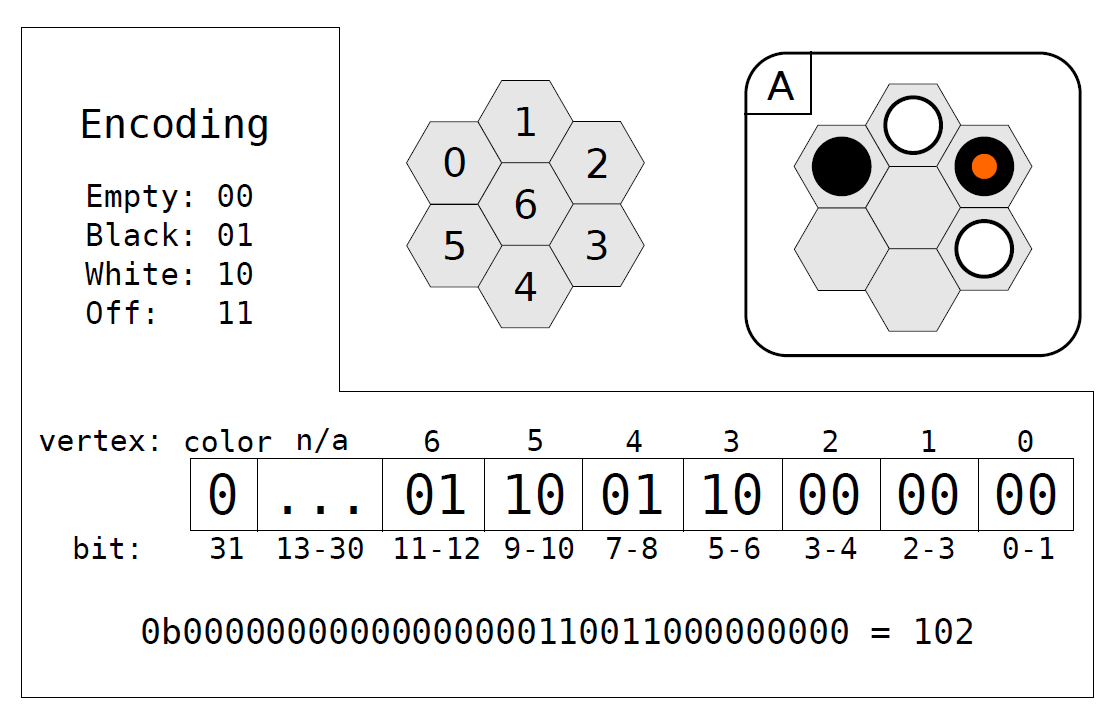}
\caption{Example of a board pattern encoded in an individual of a population optimized by EA.~\citep{pettit2012evolutionary}.}\label{fig:patterns}
\end{figure}

\cite{pettit2012evolutionary} introduced a system called \emph{Hivemind} that learns how to play abstract strategy games on regular boards. The game of choice here is Hex. The system uses MCTS/UCT and Evolution Strategies to optimize the default policy for simulations. The subject of evolution are individuals that encode:
\begin{itemize}
\item Local patterns on a board. For example, the status of a given hex cell with its 6 neighbors as shown in Figure~\ref{fig:patterns}.
\item Move selection strategies (default, uniform local, uniform local with ``tenuki'')
\end{itemize}

Thanks to using ES, the authors were able to evolve meaningful patterns that used by move selection strategies outperform baseline default policies, including uniform random selection. The obtained results are stable, i.e., the evolutionary process finds similar efficient policies each time. The authors also show that the learnt policies generalizes to different board sizes.

The tree policy, i.e., in the selection part of MCTS, can also be optimized alone. ~\cite{bravi2016evolving} investigated whether it is possible to evolve a heuristic that will perform better than UCB in the field of General Video Game Playing. They experimented with five different games from this framework: \emph{Boulderdash}, \emph{Zelda}, \emph{Missile Command}, \emph{Solar Fox} and \emph{Butterflies}. In this work, Genetic Programming (GP) is employed with a syntax tree that serves as an encoding of a chromosome. The syntax tree denotes a formula to use instead of UCB.

The possible components are:.
\begin{itemize}
\item 16 predefined constants from the range $[-30,30]$
\item Unary functions - square root, absolute value, multiplicative inverse
\item Binary functions - addition, subtraction, multiplication, division and power
\item Variables such as child visits, child depth, parent visits, child max value.
\end{itemize}

In a similar paper,~\cite{bravi2017evolving}, the authors added certain \emph{Game Variables} as possible components that describe features of the state of the game. The evolved heuristic functions perform on a similar level, i.e., not significantly better or worse, to UCB in four of five tested games. In the last game, they yield poor results. However, these heuristics give a lot of insights about particular games, e.g., whether exploration is an important factor. The authors conclude that this approach can be promising when combined with additional variables and mechanisms.

\subsection{Rolling Horizon Evolutionary Algorithm}
\label{sec:rhea}
In this subsection, we will provide a short background of Rolling Horizon Evolutionary Algorithm (RHEA). In general, RHEA is a competitor technique for creating decision-making AI agents in games. As the name implies, it is a representative of a broad class of evolutionary algorithms (EA). However, it should not be mistaken with Evolutionary-MCTS, which has been inspired by RHEA. Although MCTS and RHEA are different techniques, they are often compared in terms of performance in particular games as well combined within the same agent.

The first application of RHEA was published in \cite{perez2013rolling} for the task of navigation in single-player real-time games. For a recent summary of this method and its modifications so far, we recommend the article by~\cite{gaina2021rolling}.

Instead of doing the traditional selection and simulation steps, the RHEA algorithms evolve sequences of actions starting from the current state up to some horizon of $N$ actions. Each individual in a population encodes one sequence. Only the first action from the best sequence is played in the actual game. To compute the fitness value, the simulator applies encoded actions one by one. If an action is not legal, then it is replaced by a default one (``do nothing''). After applying the last action, the score of the game is computed as shown in Equation~\ref{eq:rhea_score}:
\begin{equation}
\label{eq:rhea_score}
h(s) = \left \{\ 
\begin{array}{ll}
	1 & \mbox{win = True},\\
	0 & \mbox{win = False}, \\
        score \in [0,1] & \mbox{otherwise}.
\end{array}
\right.
\end{equation}

The algorithm incorporates EA operators of crossover and mutation. The selection is typically the tournament selection. Elitism, i.e., directed promotion of the most fit parents to the generation, can be incorporated. In the mentioned article by~\cite{gaina2021rolling}, you can find a description of an RHEA framework tested using a set of 20 different games. As this survey is focused on modifications of the vanilla MCTS algorithm, the work by~\cite{gaina2017rolling} presents enhancements to the vanilla RHEA approach. The online self-adaptation of MCTS (e.g.) has also its counterparts for RHEA as described by~\cite{gaina2020self}.

\cite{horn2016mcts} consider a few hybridization schemes of the MCTS and EA. One is called RHEA with rollouts, in which after the application of a sequence of actions stored in a genome, a number of rollouts (Monte Carlo simulations) is performed. The final fitness is the average of the classic RHEA fitness and the average value from additional rollouts. Another hybridization scheme proposed is an ensemble approach, which first runs RHEA for a fixed time and then MCTS to check for alternative paths.

The analysis by~\cite{gaina2017analysis} compares the MCTS and RHEA approaches in General Video Game Playing. In addition, in this works, the impact of various parameters of RHEA is measured. Within the considered corpus of various games used in GVGP, there are cases, in which one method is stronger than the other. The conclusion is that neither algorithm is inherently stronger and their efficacy depends on a particular game.

\subsection{Evolutionary MCTS}
\label{sec:emcts}

\begin{figure}[!htb]
\centering
\includegraphics[width=4.6in]{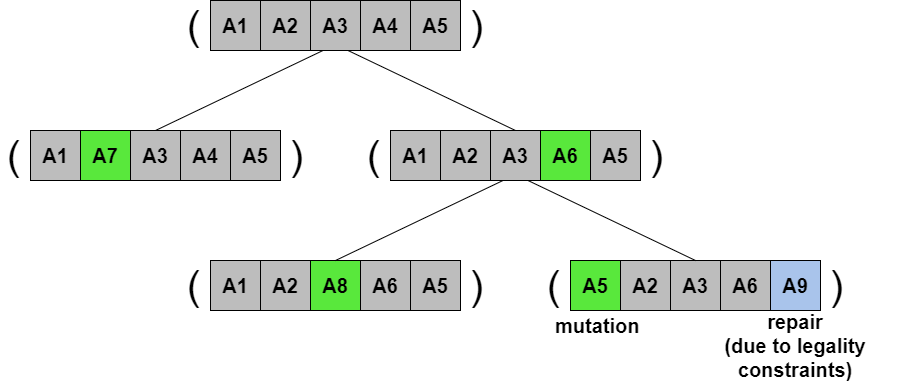}
\caption{Evolutionary MCTS. Instead of performing rollouts, sequences of actions are mutated. Repair operators are introduced in case of possible illegal moves generated this way. The figure was inspired by the work~\citep{baier2018evolutionary}. }\label{fig:rhea}
\end{figure}

The common element of the three approaches described in papers by~\cite{baier2018evolutionary} and \cite{horn2016mcts} is that EA is responsible for performing simulations.
\cite{baier2018evolutionary} propose \emph{Evolutionary MCTS} (EMCTS), which builds a different type of tree than the vanilla MCTS. It does not start from scratch, but instead all nodes, even the root, contain sequences of actions of certain length. In the presented study with \emph{Hero Academy} game, the sequences are 5 action long. Each node is a sequence, whereas edges denote mutation operators as shown in Figure~\ref{fig:rhea}. Instead of performing rollouts, the sequences in leaf nodes are evaluated using a heuristic function.


\section{MCTS Applications Beyond Games}
\label{sec:non-games}

\subsection{Planning}
\label{sec:planning}

Automated planning is one of the major domains of application of the MCTS algorithm outside games. The planning problem is typically formulated as \emph{MDP}, which was defined in Section~\ref{sec:classic_mcts}, or \emph{POMDP} (Partially Observable MDP). Similarly to games, in AI planning, there is a simulated model that can be reasoned in. The model consists of an environment with the initial state, the goal states (to achieve) and available actions. The solution is a strategy - either deterministic or stochastic, depending on a particular problem, that transitions the initial state to the goal state, playing by the rules of the environment, in the most efficient way. The most efficient fashion may be, e.g., the shortest transition or having the smallest cost. Particular applications differ between each other with respect to various constraints, extensions and assumptions. In this subsection, we focus on general planning problems. Scheduling, transportation problems and combinatorial optimization have their own dedicated sections.

\cite{vallati20152014} present a summary of the International Planning Competitions up to 2014, in which we can read that winning approaches of the probabilistic track (\textit{International Probabilistic Planning Competition}) in 2009-2014 were using Monte Carlo Tree Search. It was shown that MCTS-based planners were very strong but they lost ground in the largest problems. They often needed to be combined with other models such as Bayesian Reinforcement Learning~\citep{sharma2019robust}. This is along the motivation behind this survey, which focuses on extensions aimed at tackling the most difficult problems from the combinatorial point of view. 

Scheduling is a similar problem to planning, wherein resources are assigned to certain tasks that are performed according to a schedule. \cite{ghallab2004automated} point out that both automated planning and scheduling are often denoted as AI planning. There have been several standard implementations of MCTS to scheduling such as of~\cite{amer2013monte} for activity recognition and by~\cite{neto2020multi} for forest harvest scheduling.

MCTS has been used in other discrete combinatorial problems in a relatively standard variant, e.g. by~\cite{KUIPERS20132391} for optimizing the Horner's method of evaluating polynomials. In the work by~\cite{jia2020ultra}, MCTS is used for optimizing low latency communication. The authors present a parallelized variant, in which each parallel node is initialized with a different seed. The master processor merges the MCTS trees constructed with different seeds.~\cite{shi2020addressing} apply MCTS to automated design of generating large scale floor plans with adjacency constraints. The only non-standard modification there is discarding the rollout phase entirely. Instead the tree is expanded by each node encountered by the tree policy. The MCTS algorithm has also been used as part of a more general framework, e.g., to the imbalance subproblem in a complex framework based on cooperative co-evolution to tackle large optimization problems.

\subsubsection{Simplifications of a Problem / Model}

To tackle large planning problems,~\cite{Anand2015ASAPUCTAO} propose an abstraction of state-action pairs for UCT. Later,~\cite{anand2016oga} came up with a refined version of this idea called ``On-the-go abstractions'' . The abstraction is based on the equivalence function that maps state-action pairs to equivalent classes. In their work, \cite{Anand2015ASAPUCTAO} consider the following problems:
\begin{itemize}
\item How to compute abstractions - the algorithm is presented.
\item When to compute abstractions - time units are introduced and when the UCT is sampled up to a certain level, abstractions are computed every $t$ units, where $t$ is computed using a dynamic time allocation strategy.
\item Efficient implementation of computing abstractions - let $n$ denote the number of State-Action Pairs (SAPs). The authors reduced the complexity of the naive algorithm of computing the abstractions from $O(n^2)$ to $O(rk^2)$, where $r$ is the number of hash buckets and $k$ is the size of a bucket and $n^2 > rk^2$.
\end{itemize}

The novel method with abstractions, called ASAP-UCT, has been compared with the vanilla UCT planner and other three algorithms in six problems from three domains. The ASAP-UCT was the clear winner with statistically significantly better results in all but one comparison.

\cite{keller2012prost} introduce a system called PROST aimed at probabilistic planning problem and acting under uncertainty. As previously mentioned, in this subproblem of planning, MCTS approaches has been the most successful. PROST is equipped with several modifications of vanilla MCTS/UCT approach such as:
\begin{itemize}
\item Action pruning - the model includes the concepts of \emph{dominance} and \emph{equivalence} of actions. Dominated actions are excluded, whereas only one action from equivalent sets is considered.
\item Search space modification - the tree contains \emph{chance nodes} and \emph{decision nodes} and even after action pruning the search space is strongly connected.
\item Q-value Initialization - PROST uses \emph{single-outcome determinization} of the MDP using \emph{iterative deepening search} (IDS) to initialize Q-values and avoid the cold start.
\item Search depth limitation - search depth limit is introduced as a parameter $\lambda$.
\item Reward Locks - states, in which no matter what the agent does, it will receive the same reward. The authors explain this is often the case close to goals or dead ends. The method of detecting reward locks is presented in the paper. The MCTS uses such information in two ways: (1) to terminate a rollout earlier and (2) minimize the error induced by limited depth search by changing the simulation horizon.
\end{itemize}

\subsubsection{Modifications of UCT}

\cite{painter2020convex} modify the standard UCT formula (cf. Eq.~\ref{eq:uct}) to include the so-called \emph{Contextual Regret} and adapted to multi-objective settings. The approach has been coined as \textit{Convex Hull Monte-Carlo Tree Search}. In a multi-objective setting the reward function is multi-dimensional and produces vectors instead of single values. The approach was validated in experiments using the \textit{Generalized Deep Sea Treasure} (GDST) problem in both online and offline planning scenarios. A similar type of modification of UCT in planning problems, that is based on simple regret optimization, can be found in the article published by~\cite{feldman2014simple}

\cite{hennes2015interplanetary} consider the problem of interplanetary trajectory planning and introduces a handful of modifications to MCTS that are unique to this problem. One of the modifications, however, is the fifth step of the MCTS routine called by the authors \textit{contraction}. In the considered problem, the search-tree is often broad, i.e. the branching factor is as high as 500, but shallow, i.e., the depth is limited. In the contraption phase, the algorithm checks whether there is a sub-tree with terminal states. If each nodes of such a subtree have already been visited it can be considered solved.

\subsubsection{Hierarchical and Distributed MCTS}

\begin{figure}[!htb]
\centering
\includegraphics[width=0.9\linewidth]{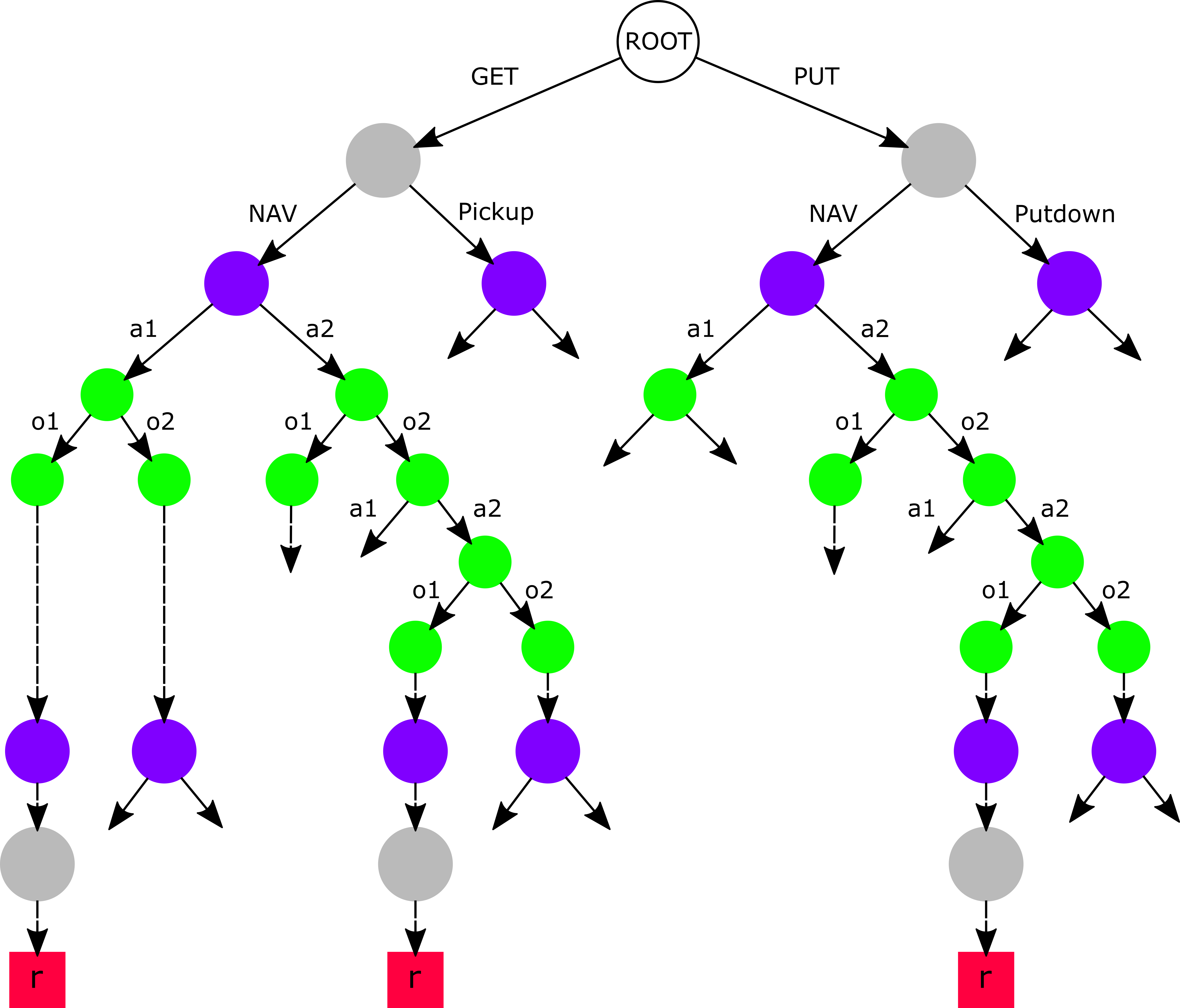}
\caption{Hierarchical Monte-Carlo Planning. The figure is inspired by~\citep{vien2015hierarchical}.} 
\label{fig:hierarchical_planning}
\end{figure}

\cite{vien2015hierarchical} propose a hierarchical Monte-Carlo planning method to tackle the problem of large search spaces and combinatorial explosion in planning. The hierarchy is introduced at the action level. The top-level tasks are subsequently decomposed into subtasks that cannot be decomposed any further. The authors discuss the concept of a hierarchical policy in POMDPs. 
This research contains a successful application of macro actions in a non-game domain. The idea is illustrated in Figure~\ref{fig:hierarchical_planning}, in which there are four actions: $\{a_{1}, a_{2}, Pickup, Putdown\}$ and three macro actions $\{GET, PUT, NAV\}$. 

The vertical dotted lines denote rollouts, whereas solid ones are tree policy executions. The green nodes denote subtree with only primitive actions. Nodes with other colors denote subtrees with particular macro actions. Each macro action defines a particular subgoal for the algorithm and has a nested routine (subroutine) for action selection. The authors show how to traverse such a MCTS tree with nodes of various hierarchy and how to expand it. The paper contains not only empirical evaluation using the Taxi problem but also theoretical results such as estimation of the bias induced by the Hierarchical UCT (H-UCT). They conclude that the proposed approach aims to mitigate two challenging problems by means of the ``curse of dimensionality'' and the ``curse of history''. Hierarchical MCTS planning is efficient whenever a hierarchy can be found in the problem. It has been featured in recent works as well~\citep{patra2020integrating}.

\subsubsection{Planning in Robotics}

\begin{figure}[!htb]
\centering
\includegraphics[width=0.8\linewidth]{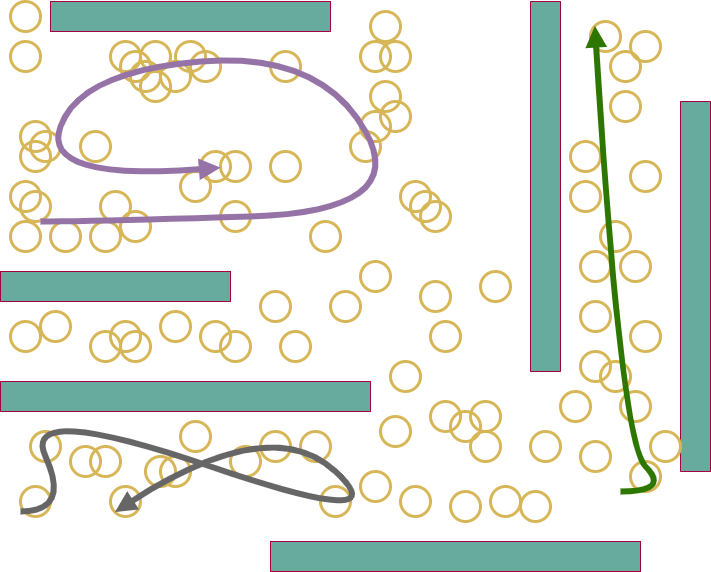}
\caption{Decentralized planning. The figure is drawn inspired by the work of~\cite{best2019dec}.} \label{fig:decentralized}
\end{figure}

MCTS has also been used for planning in robotics. \cite{best2019dec} propose a decentralized variant of MCTS for multi-robot scenario, in which each robot maintains its individual search tree. The trees in compressed forms are periodically sent to other robots, which results in an update of joint distribution over the policy space. The approach is suitable for online replanning in real-time environment. The algorithm was compared to standard MCTS with a single tree with actions of all robots included. Therefore, the action space of the tree was significantly larger than any of the decentralized trees of the Dec-MCTS approach. Apart from theoretical analysis, the experiments were carried out in 8 simulated environments based on the team orienteering problem as shown in Figure~\ref{fig:decentralized}. The Dec-MCTS achieved a median 7\% better score than the standard algorithm. 

\cite{WRJijcai15} introduce the \emph{k-Rejection Multi-Agent Markov Decision Process} $k$-RMMDP model in the context of real-world disaster emergency responders. In this model, plans can be rejected under certain conditions e.g. if the human responders decide that they are too difficult. Because their approach is based on the two-pass UCT planning as shown in Algorithm~\ref{alg:two_pass_uct}, it can be considered hierarchical as well.


\begin{algorithm}[ht]
\DontPrintSemicolon
\KwIn{The $k$-RMMDP Model: $M$, The Initial State: $(s^{0}, 0)$}
\SetKwBlock{Solve}{Solve the underlying MMDP of $M$ and build a search tree:}{}
\SetKwBlock{Create}{Create a sub MMDP:}{}
\SetKwBlock{Run}{Run UCT on the sub MMDP:}{}
\tcp{The first pass:}
\tcp{Compute the policy for $\forall s, (s, 0)$}
\Solve{
	Run UCT on the underlying MMDP: $\langle S, \overrightarrow{A}, T, R \rangle$\;
	Update all the $Q$-values: $Q(s, \overrightarrow{a}) \leftarrow Q(s, \overrightarrow{a}) \cdot \Delta_{0}(s, \overrightarrow{a})$\;
}
\tcp{The second pass:}
\tcp{Compute policy for $\forall s, j \neq 0, (s, j)$}
\ForEach{state node $(s, 0)$ of the tree}{
	\Create{
		Initial state: $(s, 0)$\;
		Normal states: $\forall j, (s, j)$\;
		Terminal states: $\forall s', (s', 0)$\;
		Actions: $\overrightarrow{A}$, Transition: $\mathbf{T}$, Reward: $\mathbf{R}$\;
	}
	\Run{
		Build a subtree with states $\forall j, (s, j)$\;
		Propagate values of $\forall s', (s', 0)$ to the subtree\;
		Propagate values of the subtree to node $(s, 0)$\;
	}	
	Update all the $Q$-values of $(s, 0)$'s ancestors in the tree\;
}
\tcp{The policy: $\pi(s^{0}, j)=argmax_{\overrightarrow{a}}Q((s^{0}, j), \overrightarrow{a})$}
\Return{the policy computed with $Q$-values in the search tree}\;
\caption{Two-Pass UCT Planning \citep{WRJijcai15}}
\label{alg:two_pass_uct}
\end{algorithm}

Monte-Carlo Discrepancy Search~\citep{clary2018monte} is another MCTS variant with applications in robotics. The idea is to find and store the current best path out to a receding horizon. Such a path is invalidated either by a change in the horizon or by stochastic factors. The simulations are performed around the best plan from the previous iteration. Then a so-called discrepancy generator selects a new node to branch from in the tree. 
The method was applied to planning the movement of a self-stable blind-walking robot in a 3D environment, which involved navigating around the obstacles.  

\subsection{Security}\label{sec:security}
One of very recent areas of MCTS application are Security Games (SG). In short, in SG methods from game theory (GT) are used to find optimal patrolling schedules for security forces (e.g. police, secret service, security guards, etc.) when protecting a set of potential targets against the attackers (criminals, terrorists, smugglers, etc.). Recently, SGs gained momentum due to rising terrorist threats in many places in the world. While the most noticeable applications refer to homeland security, e.g. patrolling LAX airport~\citep{jain2010software} or patrolling US coast~\citep{an2013deployed}, SGs can also be applied to other scenarios, for instance poaching preventing~\citep{fang2015security,gholami2018adversary,bondi2020} or cybersecurity~\citep{Kaietal2020scalable}.

In SG there are two players, the \emph{leader} (a.k.a. the \emph{defender}) who represents the security forces, and the \emph{follower} (a.k.a. the \emph{attacker}) who represents their opponents. The vast majority of SGs follow the so-called Stackelberg Game (StG), which is an asymmetric game model rooted in GT. In StG each of the two players commits to a certain \emph{mixed} strategy, i.e. a probability distribution of \emph{pure} strategies. In a typical patrolling example, a pure strategy would be a deterministic patrolling schedule which happens with certain probability. A set of such deterministic schedules (pure strategies) whose probabilities sum up to $1$ constitutes mixed leader's strategy.

The aforementioned information asymmetry of StG consists in the underlying assumption that first the leader commits to a certain mixed strategy, and only then the follower (being aware of that strategy) commits to their strategy. This assumption reflects common real-life situations in which the attacker (e.g. a terrorist or a poacher) is able to observe the defender's behavior (police or rangers) so as to infer their patrolling schedules, before committing to an attack. SGs relying on StG model are referred to as Stackelberg Security Games (SSG). SSG are generally imperfect-information games as certain aspects of the current game state are not observable by the players. Solving SSG consists in finding the Stackelberg Equilibrium (SE), which is typically approached by solving the Mixed Integer-Linear Program (MILP) corresponding to a given SSG instance~\citep{paruchuri2008playing}.

Calculation of SE is computationally expensive in terms of both CPU time and memory usage, and therefore, only relatively simple SSG models can be solved (exactly) in practice. For this reason, in many papers various simplifications are introduced to MILP SG formulations, so as to scale better for larger games, at the expense of obtaining approximate solutions. However, these approximate approaches still suffer from certain limitations of the baseline MILP formulation, e.g. the requirement of matrix-based game representation. Consequently, the vast majority of  deployed solutions consider only single-act (one-step) games.

The lack of efficient methods for solving multi-act games motivated the research on entirely different approach~\citep{KarwowskiMandziukICAISC2015} which, instead of solving the MILP, relies on massive Monte-Carlo simulations in the process of finding efficient SE approximation. 

In the next couple of years the initial approach was further developed into a formal solution method, called Mixed-UCT ~\citep{KarwowskiMandziukECAI2016}
which is generally suitable for a vast range of multi-act games.
Mixed-UCT consists in iterative computation of the leader's strategy, based on the results of MC simulations of the game, played against a gradually improving follower. The currently considered follower is defined as a combination of a certain number of past followers from previous Mixed-UCT iterations. The method scales very well in both time and memory and provides efficient SE approximations (approximate mixed leader's strategies) for various types of SGs~\citep{KarwowskiMandziukEJOR2019}.
A general concept of Mixed-UCT is depicted in Figure~\ref{security}. 

Recently, the same authors have proposed another UCT-based method for SE approximation in general-sum multi-step SGs. The method, called Double Oracle UCT (O2UCT), iteratively interleaves the following two phases: (1) guided MCTS sampling of the attacker's strategy space; and (2) constructing the defender's mixed-strategy, for which the sampled attacker's strategy (1) is an optimal response~\citep{KarwowskiMandziukAAMAS2019,KarwowskiMandziukAAAI2020}.

One of the underlying assumptions of Stackelberg Games (SGs) is perfect rationality of the players. This assumption may not always hold in real-life scenarios in which the (human) attacker's decisions may be affected by the presence of stress or cognitive biases. There are several psychological models of bounded rationality of human decision-makers, among which the Anchoring Theory (AT)~\citep{tversky1974judgment} is one of the most popular. In short, AT claims that humans have a tendency to flatten probabilities of available options, i.e. they perceive a distribution of these probabilities as being closer to the uniform distribution than it really is.
An extension of O2UCT method that incorporates the AT bias into the attacker's decision process (AT-O2UCT) has been recently considered in~\citep{KarwowskietalAAMAS2020}.  


\begin{algorithm}[ht]
\DontPrintSemicolon
\KwIn{problem specification}
\KwOut{defender's strategy - the best of strategies developed in all iterations}
$tree \leftarrow \emptyset$\;
$evaders \leftarrow [InitES()]$ \tcp{Vector of attacker's strategies in subsequent iterations initialized with the optimal strategy against uniform defender}
\While{not EndConditions()}{
	$moves \leftarrow []$\;
	\For{depth $\leftarrow$ 1...T}{
		$tree \leftarrow I2Uct(tree, EvaderStrategy(evaders), depth, moves)$\;
		$tree \leftarrow Append(moves, BestMove(tree, depth))$\;
	}
	$defender \leftarrow StrategyFromTree(tree)$\;
	$evaders \leftarrow Append(evaders, OptEvader(defender))$\;
}
\Return{the best strategy among all calculated in the loop}\;
\caption{Mixed-UCT for security games \citep{KarwowskiMandziukEJOR2019}}
\label{security}
\end{algorithm}

\subsection{Chemical Synthesis}
\label{sec:chemical_synthesis}

\cite{segler2018planning} in an article published in \emph{Nature} show an outstanding approach to planning chemical syntheses that combines MCTS with deep neural networks. These networks have been trained on most reactions known from the history of organic chemistry research. The particular type of synthesis considered in this paper is computer-aided retrosynthesis, in which molecules are transformed into increasingly simpler precursors. In this approach, named \textbf{3N-MCTS}, three neural networks guide the MCTS algorithm:
\begin{itemize}
\item Rollout policy network - trained on 17134 rules describing reactions, which were selected from the set of all published rules with a requirement for each rule to appear at least 50 times before 2015. The actions in the simulation phase of MCTS are sampled from this network.
\item In-score filter network - this network provides retroanalysis in the expansion phase and predicts whether a particular transformation resulting by applying an action from the rollout policy is expected to actually work. If not - the action is discarded and a different one has to be chosen.~\cite{awais2018mcts} based their work on a similar idea for synthesis of approximate circuits that discards circuit designs which are not promising in terms of quality. This way, the combinatorial complexity could be drastically reduced.
\item Expansion policy network - this network selects the most promising node to add in the MCTS tree. This is a significant change compared to the standard approach, in which the first non-tested action defines the node to expand.
\end{itemize}
Combining deep neural networks with MCTS has been proven a viable approach to various chemical or physical problems including \emph{FluidStructure Topology Optimization} by~\cite{gaymann2019deep}. The problem is formulated as a game on a grid, similar to Go.

\subsection{Scheduling}
\label{sec:scheduling}

Combining MCTS with additional heuristics is a pretty common approach to large computational problems.~\cite{asta2016combining} use MCTS to generate initial solutions to a Multi-mode Resource-constrained Multi-project Scheduling Problem, which are then modified through a local search guided by carefully crafted hyper-heuristics. 

In order to create a test-bed for Computational
Intelligence (CI) methods dealing with complex, non-deterministic
and dynamic environments \cite{WaledziketalSSCI2014} introduces a new class of problems, based on the real-world task of project scheduling and executing with risk management. They propose Risk-
Aware Project Scheduling Problem (RAPSP) as a significant
modification of the Resource-Constrained Project Scheduling
Problem (RCPSP).

In \citep{waledzik2015risk}, they describe a method called \emph{Proactive UCT} (ProUCT) that combines running heuristic solvers with the UCT algorithm. The heuristic solvers create baseline schedules, whereas UCT is responsible for testing various alternative branches caused by dynamic events. The method is further extended and summarized in \citep{WaledzikMandziukINFSCI2018}.

\cite{wijaya2013effective} present an approach to energy consumption scheduling in a smart grid system that is based on MCTS. The application of the algorithm is relatively standard, however, the solutions can vary in terms of the so-called \emph{laziness}. The lazier the solution, the less frequent consumption schedule changes. Since the lazier solutions are preferable, the UCT formula includes an additional penalty that is inversely proportional to the laziness value of the solution in a node.
 
An interesting scheduling approach inspired by AlphaGo~\cite{silver2016mastering} has been proposed in~\cite{8885307}. Spear system, briefly summarized in section~\ref{sec:AlphaGo}, was reported to outperform modern heuristics and reduce the maskespan of production workloads by 20\%.

\subsection{Vehicle Routing}
\label{sec:vehicle_routing}

Transport problems such as Capacitated Vehicle Routing Problem (CVRP) are special class of problems that often involve both planning and scheduling. We recommend the surveys 
by~\cite{caceres2014rich} or~\cite{MandziukVRPSurvey2019} for the introduction to various problem specifications, the most commonly applied techniques other than MCTS and the latest advances. There are many variants of CVRPs. The common elements include planning routes for vehicles, which resembles solving multiple \emph{travelling salesman problems} (TSP), as well as allocating goods to the vehicles that need to be delivered to customers, which resembles the \emph{knapsack} problem. Therefore, even the basic version of CVRP is a cross of two NP-hard problems. The particular CVRP variants may operate with discrete or continuous time, time windows for loading and unloading, traffic, heterogeneous goods, specific client demands, the number of depots the vehicle start in, pickup and delivery, split delivery and many more. The goal, in the order of importance, is to: (1) deliver goods to the clients that meet their demand, (2) minimize the total length of the routes, (3) sometimes minimize the number of vehicles used. At the same time, all the constraints such as not exceeding the capacity of any vehicle must be preserved. 

There have been several papers, in which an MCTS-based approach has been applied to a CVRP problem or a similar one such as Multi-Objective Physical Travelling Salesman Problem. Since we focus on MCTS in this survey and the approaches share many similarities regarding how this algorithm is applied, we grouped them into topics:

\paragraph{Problem simplification by having a ``good'' initial solution} - featured in the articles by~\cite{powley2013monte,mandziuk2017uct,mandziuk2016simulation,juan2013using}. ~\cite{powley2013monte} use a classical TSP solver for the initial solution and MCTS acted as a steering controller over it. ~\cite{mandziuk2017uct,mandziuk2016simulation} as well as~\cite{juan2013using} generate the initial solution with the use of a modified Clarke and Wright savings algorithm. In the mentioned works by~\cite{mandziuk2017uct,mandziuk2016simulation}, the MCTS algorithm was used to test various incremental modifications of the initial solutions. Therefore, an action of the MCTS algorithm resembled a mutation operator in the evolutionary algorithm. \cite{juan2013using} employ the MCTS algorithm to generate a distribution of various starting solutions. Each solution is sent to a parallel processing node for further modification.

\paragraph{Problem factorization} - featured in papers by~\cite{kurzer2018decentralized}, as well as \cite{mandziuk2017uct, mandziuk2016simulation}, \cite{Mandziukbook2018}. Here instead of maintaining one MCTS tree for the complete solution, each vehicle (route) is attributed with a separate tree that MCTS iterates over. However, when the simulation takes into account effects of applying actions in each tree in a synchronized manner. The final decision, i.e., ``the action to play'' is a vector of actions from each tree.
\paragraph{Macro-actions and action heuristics} - featured in~\cite{powley2013monte,Mandziukbook2018,mandziuk2017uct,mandziuk2016simulation,kurzer2018decentralized}. In each approach, the authors proposed actions that are not the atomic actions available in the model. This modification was commonly referred to in this survey as it drastically reduces the search space. \cite{kurzer2018decentralized} interleave macro actions with regular actions.
\paragraph{Replacement of the UCT formula} - featured in~\cite{mandziuk2015uct,kurzer2018decentralized}. In the latter work the UCT formula is replaced by a solution assessment formula designed for the CVRP problem to better distinguish between good and poor routes. Three gradation patterns were tested - hyperbolic, linear and parabolic. \cite{kurzer2018decentralized} replace the UCT formula by the $\epsilon-Greedy$ strategy.
\paragraph{Depth limited search} - featured in~\cite{powley2013monte}. The fitness function was introduced to evaluate non-terminal states.

\subsection{Multi-Domain MCTS with Heuristics}
\cite{sabar2015population} present a relatively general framework that combines MCTS with heuristics. Herein, MCTS tree is composed of nodes that correspond to low level heuristics applied to given states. Therefore, the whole approach can be considered as a hierarchical hyper- or meta-heuristic. The feasibility of the approach was tested across six different domains. The results were competitive with the state-of-the-art methods. 

\section{Parallelization}
\label{sec:parallelization}

The MCTS algorithm consists of iterations shown in Fig.~\ref{fig:mcts}. The iterations as well as the respective phases within an iteration are performed sequentially in the classical variant. However, there have been many approaches to parallelization of the algorithm predominantly aimed at increasing the overall number of iterations per second. The more statistics gathered, the more confident the MCTS/UCT evaluation. In many traditional approaches to the parallel execution of algorithms such as matrix multiplication ones, the result of sequential and parallel versions must be exactly the same. However, in contrast to them, MCTS implementations drift away from the classic version of the algorithm in order to enable parallelism. Therefore, they can be considered variants or modifications of MCTS. For instance, the action selection formula may no longer be optimal, but multiple selections can be done at the same time and, as a result, the overall performance of the algorithm might be better. 

\subsection{Base Approaches}

There have been three major approaches to parallel MCTS proposed - \emph{Leaf Parallelization}, \emph{Root Parallelization} and \emph{Tree Parallelization}.
\begin{enumerate}
\item \textbf{Leaf Parallelization}~\citep{paralleluct} - the simplest variant, in which there is one MCTS/UCT tree, which is traversed in the regular way in the selection phase. Once a new node is expanded, there are $K$ independent playouts performed in parallel instead of just one. All of them start with the same newly added node. The results of simulations are aggregated, e.g. added or averaged, and back-propagated in the regular fashion as in the sequential algorithm. In Leaf Parallelization, the tree does not grow quicker compared to the non-parallel variant. However, the evaluation of new nodes is more accurate. 
\item \textbf{Root Parallelization}~\citep{chaslot2008parallel} - maintains $K$ parallel processes. Each process builds its independent MCTS/UCT as the sequential algorithm would do. At certain synchronization points, e.g. just before making the actual (non-simulated) decision, the statistics from the first level in the tree are aggregated from the parallel processes. Typically, weighted aggregation is performed based on the number of visits. Although, this method does not result in deeper trees, the combined statistics of nodes are more confident. Moreover, Root Parallelization comes with minimal communication overhead.
\item \textbf{Tree Parallelization}~\citep{cazenave2008parallel} - like in Leaf Parallelization and in contrast to Root Parallelization, there is one MCTS/UCT tree. The tree is shared, iterated and expanded by $K$ parallel processes. In contrast to Leaf Parallelization, the whole iteration is done by such process and not only the playout. Therefore, synchronization mechanisms such as \emph{locks} and \emph{mutexes} are used. In order to avoid selecting the same node by subsequent processes, the so-called \emph{Virtual Loss} is usually used. It tells to assign the worst possible outcome for the active players as soon as the node is visited and correct this result in the backpropagation phase. Such a mechanism reduces the UCT score and discourages the algorithm to follow the previous choices until the correct result is back-propagated.  
\end{enumerate}

The algorithms that dynamically modify policies are particularly difficult to implement in parallel.~\cite{graf2015adaptive} show that manipulation of a playout policy can reduce the efficiency of parallelization. The authors use adaptive weights (c.f. Figure~\ref{adaptive_playout_policy} in Section~\ref{sec:perfect_info}) that are updated after each playout and shared among all threads. Although they lead to a stronger player, they require synchronization using a global lock which is harmful for computational performance. The performance ratio between adaptive and static playouts in case of 16 threads computations has been measured as 68\%.

\subsection{Using Specialized Hardware}

\cite{barriga2014parallel} propose a parallel UCT search that runs on modern GPUs. In their approach, the GPU constructs several independent trees that are combined using the idea of Root Parallelism. In addition, each three is assigned a multiple of 32 GPU threads for leaf parallelism, i.e., simulations done in parallel from the same node. The method was tested using the game \emph{Ataxx}. Various setups parameterized by the number of trees and threads have been tested . The best variant achieved a 77.5\% win-rate over the sequential counterpart. However, there are many challenges to such an approach. Firstly, the game simulator must be implemented on the GPU, which is often infeasible depending on the complexity of the game. Secondly, the programmer has no control over thread scheduling and CPU/GPU communication, which can make the approach risky to apply in tournament scenarios that require precise time controls.

\cite{mirsoleimani2015scaling} study the MCTS method running in parallel on Intel Xeon Phi, which is a processing unit designed for parallel computations. Each unit allows shared memory scaling up to 61 cores and 244 threads. In their paper, there are three contributions outlined: (1) the analysis of the performance of three common threading libraries on the new type of hardware (Xeon Phi), (2) a novel scheduling policy and (3) a novel parallel MCTS with grain size control. The authors report $5.6$ times speedup of the modified MCTS variant compared to the sequential version on the Xeon E5-2596 processor. The proposed variant is based on Tree Parallelization. The iterations are split into \emph{chunks}, which are tasks to be executed serially.

Finally,~\cite{Hufschmitt2015MCTSPP} present a parallelization method aimed for \emph{Multi-Purpose Processor Arrays} (MPPA). Each MPPA chip contains 256 processing cores organized into 16 clusters. The authors investigate parallelization of various elements and their impact on the overall performance of the resulting player. Firstly, distributed propnet evaluation is proposed. The propnet is an inference engine responsible for game-state manipulations, i.e., computing the next state in the simulation. This type of parallelization ended up introducing too much of the synchronization overhead. Secondly, playouts parallelization was evaluated requiring fewer synchronization layers. The authors report ``good'' scaling in three games (TicTacToe, Breakthrough and EightPuzzle)  with the number of clusters ranging from 1 to 16 and the maximum number of 16 threads per cluster. However, relatively high synchronization overhead was still reported and authors conclude that there are many ways, in which the approach could be further improved.

\subsection{Lock-Free Parallelization}

A ``lock'' is a mechanism that allows to give a thread the exclusive access to certain code section. It is to prevent deadlocks, races, crashes and inconsistent memory state. In the MCTS algorithm, there are two critical moments that are usually placed within the lock - expansion of a new node and back-propagation. The shared selection phase can also be a potential race condition scenario depending on the implementation.~\cite{mirsoleimani2018lock} propose a novel lock-free algorithm that takes advantage of the C++ multi-threading-aware memory model and the \emph{atomic} keyword. The authors shared their work as an open source library that contains various parallelization methods implemented in the lock-free fashion. The efficacy was shown using the game of Hex and both Tree Parallelization and Root Parallelization.

\subsection{Root-Tree Parallelization}

The approach proposed by~\cite{swiechowski2016hybrid} defines a way of combining Root- and Tree Parallelization algorithms that leads to a very high scalability. The setup consists of the so-called worker nodes, hub nodes and the master node, for the purpose of message passing and scalability. Each worker node maintains its own MCTS/UCT tree that is expanded using Leaf Parallelization. On top of them, there is a Root Parallelization method (running on hub and master nodes) that gathers and aggregates statistics from worker nodes in a hierarchical manner. In addition, the authors propose a novel way, called \emph{Limited Root-Tree Parallelization}, of splitting the tree into certain subtrees to be searched by worker nodes. The subtrees are not entirely disjoint as some level of redundancy is desirable for Root Parallelization. This allows to push the scaling boundary further when compared with just the hierarchical combination of Root and Tree Parallelization methods. The resulting hybrid approach is shown to be more effective than either method applied separately in a corpus of nine General Game Playing games.

\subsection{Game Specific Parallelization}

Parallelism can become an enabler to solving games, which are not too combinatorially complex. One such example is Hex~\citep{mcts-hex}.~\cite{liang2015job} propose an approach to solving Hex in a parallel fashion. The work builds upon the Scalable Parallel Depth-First Proof-Number Search (SPDFPN) algorithm, which has the limitation that the maximum number of threads that can be utilized cannot be greater than the number of CPU cores. The proposed Job-Level UCT search no longer has this limitation. The authors introduced various techniques aimed at optimizing the workload sharing and communication between the threads. The resulting solver is able to solve 4 openings faster than the previous state-of-the-art approach. However, solving the game universally is not yet feasible.

\cite{schaefers2014distributed} introduce a novel approach to parallelization of MCTS in Go. The novelty includes efficient parallel transposition table and dedicated compute nodes that support broadcast operations. The algorithm was termed \emph{UCT-Treesplit} and involves ranking nodes, load balancing and certain amount of node duplication (sharing). It is targeted at HPC clusters connected by \emph{Infiniband}. The MPI library is employed for message passing. The major advantage of using the \emph{UCT-Treesplit} algorithm is its great scalability and rich configuration options that enable to optimize the algorithm for the given distributed environment.

\subsection{Information Set Parallelization}

\cite{sephton2014parallelization} report a study aimed at testing the base parallelization methods, i.e. Leaf, Root and Tree, adapted to Information Set MCTS (ISMCTS). In addition, the Tree Parallelization was included in two variants - with and without the so-called \textit{Virtual Loss}. The game of choice for the experiments is a strategic card game \emph{Lords of War}. The authors conclude that the Root Parallelization method is the most efficient approach when combined with ISMCTS.

\section{Conclusions}
\label{sec:conclusions}

\label{sec:tables}

\begin{table*}[htbp]
\begin{footnotesize}
\caption{References grouped by application}
\label{tbl:applications}
\begin{tabular}{|l|l|}
\hline 
\textbf{Games} &\textbf{References} \\
\hline 
\hline 
Go&  \makecell[l]{\cite{Silver1140}, \cite{silver2016mastering}, \cite{silver2017mastering}, \\ \cite{browne2012problem}, \cite{graf2016adaptive}, \\ \cite{graf2015adaptive}, \cite{chen2012dynamic}, \cite{ikeda2013production},\\ \cite{baier2015time},  \cite{graf2014common},\\ \cite{yang2020learning}, \cite{wu2018multilabeled}, \\  \cite{brugmann1993monte}, \cite{coulom2007computing},
\cite{drake2007move}}  \\
\hline 
Hex &  \cite{chao}, \makecell[l]{\cite{gao2017move}, \cite{takada2019reinforcement}  \\
\cite{huang2013mohex}, \cite{gao2019hex}} 
\\
                    
\hline 
\makecell[l]{Other with perfect \\information} & \makecell[l]{\cite{chang2018big}, \cite{tang2016adp}, \cite{zhuang2015improving},\\ \cite{soemers2019learning}, \cite{benbassat2013evomcts}}\\
\hline 
Card games & \makecell[l]{\cite{cowling2012ensemble}, \cite{santos2017experiments}, \cite{maciek2018improving}, \\ \cite{zhang2017improving}, \cite{sephton2014lordsofwar}, \cite{sephton2015experimental}, \\ \cite{baier2018emulating}, \cite{ihara2018pokemon}, \cite{di2018traditional},\\ \cite{choe2019enhancing}, \cite{santos2017experiments}, \cite{godlewski2021optimisation}, \\ \cite{goodman2019re}}   \\ 
\hline 
Arcade Video games & \makecell[l]{\cite{alhejali2013pacman}, \cite{pepels2012enhancements}, 
\cite{pepels2014real}, \\ \cite{nguyen2012pacman}, \cite{jacobsen2014mario}, \cite{kartal2019action}, \\ \cite{guo2014deep}, \cite{pinto2018hierarchical}, \cite{ilhan2017monte}, \\
\cite{perick2012comparison}, \cite{gedda2018monte}, \cite{zhou2018hybrid}} \\
\hline 
\makecell[l]{Real Time Strategy\\ games} & \makecell[l]{\cite{uriarte2017single}, \cite{justesen2014scriptutc},  \\  \cite{moraes2018action}, \cite{uriarte2014high}, \\ \cite{yang2018learning}, \cite{ontanon2016informed}, \cite{uriarte2016improving}\\
\cite{preuss2020games}}  \\ 
\hline 
General Game Playing & \makecell[l]{\cite{WaledzikMandziukAGI2011}, \cite{swiechowski2013self},\\  
\cite{cazenave2016playout}, \cite{sironi2018self}, \cite{nelson2016investigating}, \\  \cite{finnsson2008simulation},  \cite{finnsson2010learning},\\ \cite{WaledzikMandziukTCIAIG2014}, \cite{trutman2015creating},\\  
\cite{sironi2016comparison}, \cite{SwiechowskietalTCIAIG2016},\\
\cite{sironi2019comparing}} \\
\hline
\makecell[l]{General Video \\ Game Playing} & \makecell[l]{  \cite{park2015mcts}, \cite{joppen2017informed}, \\  \cite{frydenberg2015investigating}, \cite{guerrero2017beyond}, \\ \cite{perez2014knowledge},  \cite{horn2016mcts}, \\ \cite{gaina2017analysis},  \cite{de2016monte}, \cite{soemers2016enhancements}} \\
\hline
Game development & \makecell[l]{\cite{hunicke2005case}, \cite{keehl2018monster}, \cite{keehl2019monster}, \\ \cite{maia2017using}, \cite{baier2018emulating}, \cite{zook2019monte}, \\
\cite{demediuk2017monte}, \cite{ishihara2018monte}, 
\cite{khalifa2016modifying} }\\
\hline
\hline
\textbf{Non-games} &\textbf{References}\\
\hline
Optimisation & \makecell[l]{\cite{uriarte2016improving}, \cite{KUIPERS20132391}, \cite{jia2020ultra}, \\ \cite{shi2020addressing}, \cite{segler2018planning}, \cite{awais2018mcts}, \\ \cite{gaymann2019deep}, \cite{sabar2015population}}\\
\hline
\makecell[l]{Security}  &  \makecell[l]{\cite{KarwowskiMandziukICAISC2015}, 
\cite{KarwowskiMandziukEJOR2019},\\ 
\cite{paruchuri2008playing},
\cite{KarwowskiMandziukAAAI2020}}\\
\hline
\makecell[l]{Chemical Synthesis}  &  \makecell[l]{
\cite{segler2018planning}, \cite{awais2018mcts}, \cite{gaymann2019deep}
}\\
\hline
Planning and Scheduling & \makecell[l]{\cite{ghallab2004automated}, \cite{Anand2015ASAPUCTAO}, \cite{WaledziketalSSCI2014},\\
\cite{anand2016oga}, \cite{keller2012prost}, \cite{painter2020convex},\\ 
\cite{feldman2014simple}, \cite{waledzik2015risk},\\ \cite{vien2015hierarchical},
\cite{patra2020integrating}, \cite{best2019dec},\\ \cite{WRJijcai15}, \cite{clary2018monte}, \cite{8885307},\\
\cite{amer2013monte}, \cite{neto2020multi}, \cite{asta2016combining},\\ \cite{wijaya2013effective},
\cite{WaledzikMandziukINFSCI2018}}\\
\hline
Vehicle Routing &\makecell[l]{ \cite{caceres2014rich}, \cite{powley2013monte}, \cite{Mandziukbook2018}\\ \cite{mandziuk2017uct}, \cite{mandziuk2015uct}, \\ \cite{mandziuk2016simulation} , \cite{juan2013using}, \cite{kurzer2018decentralized}\\
\cite{MandziukVRPSurvey2019}}\\
\hline
\end{tabular} 
\end{footnotesize}
\end{table*}


\begin{table*}[htbp]
	\begin{footnotesize}
		\caption{References grouped by method, part 1/2, continued on Table~\ref{tbl:methods2} }
		\label{tbl:methods1}
		\begin{tabular}{|l|c|l|}
			\hline 
			\textbf{Methods} & \textbf{Section} & \textbf{References} \\
			\hline 
			\hline 
			Classic MCTS & \ref{sec:classic_mcts}& \makecell[l]{
			\cite{uct}, \cite{ucb1tuned},\\
			\cite{auer2002finite}, \cite{thompson1933likelihood}, \\ \cite{bai2013bayesian}, \cite{gelly2011monte}, \\
			\cite{finnsson2010learning}, \cite{finnsson2011cadiaplayer}, \\
			\cite{tt}, \cite{hh}, \\ 
			\cite{nst}, \cite{gaudel2010principled}, \\
			\cite{chaslot2009meta}} \\
			\hline 
			Action Reduction & \ref{sec:action_reduction} & \makecell[l]{
			\cite{sephton2014lordsofwar}, \cite{justesen2014scriptutc},\\ \cite{churchill2013portfolio}, \cite{de2016monte},\\ \cite{subramanian2016efficient}, \cite{gabor2019subgoal},\\
			\cite{moraes2018action}, \cite{uriarte2014high},\\ \cite{ontanon2016informed}, \cite{baier2012beam},\\ \cite{pepels2012enhancements}, \cite{soemers2016enhancements},\\
			\cite{zhou2018hybrid}, \cite{chaslot2008progressive}, \\ \cite{gedda2018monte}, \cite{liu2015regulation}, \\ \cite{mandai2016linucb}, \cite{tak2014monte}, \\
			\cite{yee2016monte}, \cite{perick2012comparison}\\
			\cite{ikeda2013efficiency}, \cite{coulom2007computing}, \\ \cite{imagawa2015enhancements}, \cite{demediuk2017monte}
			}\\
			\hline 
			UCT Alternatives & \ref{sec:uct_alternatives}& \makecell[l]{
			\cite{brugmann1993monte}, \cite{gelly:go}, \\ \cite{sironi2016comparison}, \cite{cazenave2015generalized}, \\ \cite{sarratt2014converging}, \cite{sironi2019comparing}, \\ \cite{browne2012problem}, \cite{imagawa2015enhancements},\\ \cite{graf2016adaptive}, \cite{graf2015adaptive}, \\	
			\cite{gudmundsson2013sufficiency}, \cite{baier2014mcts}, \\ \cite{baier2013winands}}\\
			\hline 
			Early Termination & \ref{sec:early_termination}& \makecell[l]{
			\cite{lorentz2016using}, \cite{lanctot2014monte}, \\
			\cite{WaledzikMandziukTCIAIG2014},\cite{wu2018multilabeled}, \\
			\cite{goodman2019re}, \cite{cazenave2015generalized}, \\
			\cite{sironi2016comparison}}\\ 
			\hline 
			Determinization & \ref{sec:determinization}& \cite{cowling2012infoset}, \cite{cowling2012ensemble}\\ 
			\hline 
			Information Sets & \ref{sec:ISMCTS}& \makecell[l]{ \cite{cowling2012infoset}, \cite{KarwowskiMandziukICAISC2015}, \\
			\cite{wang2015belief}, \cite{cowling2015bluffing}, \\ 
			\cite{lisy2015online}, \cite{furtak2013recursive}, \\ \cite{KarwowskiMandziukEJOR2019}, \cite{ihara2018pokemon}, \\ \cite{di2018traditional}, \cite{sephton2015experimental}, \\ 	
			\cite{KarwowskiMandziukAAAI2020}, \cite{goodman2019re},\\ 
			\cite{powley2013bandits}, \cite{uriarte2017single}		
			}\\
			\hline 
			Heavy Playouts & \ref{sec:heavy}& \makecell[l]{ \cite{browne2012problem}, \cite{swiechowski2013self}, \\\cite{alhejali2013pacman},
			\cite{godlewski2021optimisation}}\\
			\hline 
			Policy Update & \ref{sec:policy}& \makecell[l]{ \cite{kao2013incentive}, \cite{cazenave2015generalized}, \cite{cazenave2016playout} \\
			\citep{trutman2015creating}, \cite{santos2017experiments},\\ \citep{maciek2018improving} }\\ 
			\hline 
			Master Combination & \ref{sec:combination}& \makecell[l]{ 
			\cite{swiechowski2020game}, \cite{nguyen2012pacman}, \\ \cite{pepels2012enhancements}, \cite{pepels2014real}, \\ \cite{jacobsen2014mario}, \cite{nelson2016investigating}, \\ \cite{soemers2016enhancements}, \cite{tak2012ngrams}, \\  \cite{frydenberg2015investigating}, \cite{guerrero2017beyond}, \\ \cite{kim2017opponent}, \cite{goodman2019re}, \\ 
            \cite{khalifa2016modifying}
			}\\
			\hline 
			Opponent Modelling & \ref{sec:opponent_modelling} & \makecell[l]{ 
			\cite{goodman2020does}, \cite{cowling2015bluffing}, \\ 
			\cite{baier2012beam}, \cite{kim2017opponent}, \\
			\cite{hunicke2005case},	\cite{nguyen2012pacman}, \\ \cite{baier2018emulating}
			}\\
			\hline
			
			\hline
		\end{tabular} 
	\end{footnotesize}
\end{table*}			
			
\begin{table*}[htbp]
	\begin{footnotesize}
		\caption{References grouped by method, part 2/2, continuation from Table~\ref{tbl:methods1} }
		\label{tbl:methods2}
		\begin{tabular}{|l|c|l|}
			\hline 
			\textbf{Methods} & \textbf{Section} & \textbf{References} \\
			\hline 			
			
			\makecell[l]{MCTS + Neural \\Networks} & \ref{sec:mcts_nn} & \makecell[l]{
			\cite{chang2018big}, \cite{chao}, \\
			\cite{gao2017move}, \cite{takada2019reinforcement}, \\ \cite{yang2020learning}, \cite{wu2018multilabeled},\\ \cite{tang2016adp},  \cite{gao2019hex} \\ \cite{zhang2017improving}, \cite{maciek2018improving}, \\ \cite{zhuang2015improving}, \cite{yang2018learning}, \\ \cite{baier2018emulating}, \cite{soemers2019learning}, \\ \cite{kartal2019action}, \cite{guo2014deep}, \\ \cite{pinto2018hierarchical}
			}\\
			\hline 
			\makecell[l]{MCTS + Reinforcement \\Learning} & \ref{sec:ml}& \makecell[l]{\cite{Silver1140}, \cite{silver2016mastering}, \cite{silver2017mastering},\\ \cite{vodopivec2014enhancing}, \cite{ilhan2017monte},\\ \cite{NIPS2017Anthonyetal}}\\ 
			\hline 
			\makecell[l]{MCTS + Evolutionary \\Algorithms} & \ref{sec:mcts_evolutionary} & \makecell[l]{
			\cite{benbassat2013evomcts}, \cite{alhejali2013pacman}, \\ \cite{lucas2014fast}, \cite{pettit2012evolutionary}, \\
			\cite{bravi2016evolving}, \cite{bravi2016evolving}, \\
			\cite{perez2013rolling}, \cite{gaina2021rolling}, \\
			\cite{baier2018evolutionary}, \cite{perez2014knowledge}, \\ \cite{horn2016mcts}, \cite{gaina2017analysis}
			}\\ 
			\hline 
			Classic Parallelization & \ref{sec:parallelization} & \makecell[l]{\cite{paralleluct}, \cite{chaslot2008parallel},\\ \cite{cazenave2008parallel}} \\
			\hline 
			New Parallelization & \ref{sec:parallelization} & \makecell[l]{\cite{graf2015adaptive}, \cite{barriga2014parallel},\\ \cite{mirsoleimani2015scaling}, \cite{Hufschmitt2015MCTSPP},\\ \cite{mirsoleimani2018lock}, \cite{swiechowski2016hybrid},\\ \cite{mcts-hex}, \cite{liang2015job},\\ \cite{schaefers2014distributed}, \cite{sephton2014parallelization}} \\ 
			\hline
		\end{tabular} 
	\end{footnotesize}
\end{table*}


MCTS is a \emph{state-of-the-art} tree-search algorithm used mainly to implement AI behavior in games, although it can be used to support decision-making processes in other domains as well.
The baseline algorithm, described in Section~\ref{sec:classic_mcts}, was formulated in 2006, and since then multitude of enhancements and extensions to its vanilla formulation have been published. Our main focus in this survey is on works that have appeared since 2012, which is the time of the last major MCTS survey authored by~\cite{mctsSurvey}. Our literature analysis yielded 240 papers cited and discussed in this review, the vast majority of which fell within the above-mentioned time range. 
An overview of the considered papers grouped by application domains and by enhancements introduced to baseline MCTS are presented in Tables~\ref{tbl:applications} and~\ref{tbl:methods1}, respectively. 

In their 2012 survey,~\cite{mctsSurvey} concluded that MCTS would be ``extensively hybridised with other search and optimisation algorithms''. In the consecutive  8 years this prediction was positively validated in many papers and, based on their analysis, the following main trends of such hybridized approaches can be distinguished:
\begin{enumerate}
\item \textbf{MCTS + ML} - combining MCTS with machine learning models, in particular deep neural networks trained with reinforcement learning \citep{silver2016mastering, NIPS2017Anthonyetal}, has become a significant research trend which is discussed in Section~\ref{sec:ml}.
Its popularity is to a large extent  due to the unprecedented success of \emph{AlphaGo} \citep{silver2016mastering} and the rapid increase of interest in the ML field, in general. Further extensive growth in this direction can be safely envisaged.
\item \textbf{AutoMCTS} - an enhancement of MCTS with self-adaptation mechanisms \citep{swiechowski2013self, gaina2020self}, which rely on finding optimal hyperparameters or the most adequate policies for a given problem being solved. As discussed in Section~\ref{sec:ml}, MCTS is often combined with optimization techniques (e.g. evolutionary approaches) that are executed \emph{offline} in order to find the set of parameters for the \emph{online} MCTS usage \citep{horn2016mcts, lucas2014fast,pettit2012evolutionary, bravi2016evolving}.
\item \textbf{MCTS + Domain Knowledge} - an inclusion of domain knowledge into MCTS operation, either in the form of the heavy playouts, or action prioritization, or exclusion of certain paths in the search space (action reduction), which are not viable or plausible due to the possessed expert knowledge. Other possibilities include building a light evaluation function for action assessment or early MCTS termination due to certain problem-related knowledge \citep{lorentz2016using}.
\end{enumerate}
Elaborating more on the last point, let us observe that MCTS has become the state-of-the-art technique for making strong computer players in turn-based combinatorial games. It has led to many breakthroughs, e.g. in Go and Chess. One of its main strengths is no need for game-related expert knowledge. In the vanilla form, only the simulation engine, i.e. rules of the game, is needed. Hence, it has been widely applied in multi-game environments such as General Game Playing and General Video Game AI. 

At the same time, embedding knowledge into MCTS without introducing too much bias is a potential way of improving its performance. 

We expect that injecting  knowledge into MCTS, especially in a semi-automatic fashion, is one of the future directions.  
We consider a direct inclusion of a human knowledge into the MCTS algorithm as a special case of domain knowledge utilization: 

\begin{enumerate}
\setcounter{enumi}{3}
\item \textbf{Human-in-the-Loop} - a direct implosion of human expert domain knowledge to the MCTS operational pattern by means of  a \emph{direct, online} steering of the MCTS by the human operator \citep{SwiechowskietalIARIA2015}. A human expert has the power of excluding parts of the search tree, disable/enable certain actions or increase the frequency of visits in certain search space states (tree search nodes). 
In this context, it may be worth trying to develop some frameworks for semi-automatic translation of expert knowledge into MCTS steering or adjusting so as to make the above Human-Machine model of cooperation as effortless and efficient as possible.
We believe that such  direct teaming of the human expert (knowledge) with (the search speed of) MCTS has potential to open new research path in various complex domains, extending beyond combinatorial games or deterministic problems.
\end{enumerate}

The fifth MCTS related research prospect that we envisage is
\begin{enumerate}
\setcounter{enumi}{4}
\item \textbf{MCTS in New Domains} - in principle, MCTS can be applied to any sequential decision-making problem that can be represented in the form of a tree whose leaf nodes (solutions) can be assessed using some form of an evaluation / reward function. This includes combinatorial optimization \citep{sabar2015population}, planning \citep{feldman2014simple}, and scheduling \citep{neto2020multi} problems encountered in a variety of fields, e.g. logistics \citep{kurzer2018decentralized}, chemistry \citep{segler2018planning}, security \citep{KarwowskiMandziukEJOR2019}, or in UAV (Unmanned Aerial Vehicle) navigation \citep{Chenetal2016UAV}. We are quite confident that MCTS will gradually become more and more popular in the areas other than combinatorial games.
\end{enumerate}

Finally, one of the potential research prospect, which we regard more as an opportunity than a definite prediction, is 
\begin{enumerate}
\setcounter{enumi}{5}
\item \textbf{MCTS in Real Time Games} - although MCTS has been applied to some real-time video games such as \emph{Pac-Man} \citep{pepels2014real} or \emph{Starcraft} \citep{uriarte2016improving}, 
it is not as popular a tool of choice in this field as it is in combinatorial games. \cite{rabin2013game} summarizes the techniques that are predominantly used in real-time video games, which include Behaviour Trees, Goal-Oriented Action Planning, Utility AI, Finite State Machines and even simple scripts. The main reason  for using them is the limited computational budget. Video games tend to have quite complex state representations and high branching factors. Moreover, the state can change rapidly in real-time. In research environment, it is often possible to allocate all available resources  to the AI algorithms. In commercial games, the AI module is allowed to use only a fraction of computational power of a single consumer-level machine. 
\end{enumerate}

Besides computational complexity, the other reason for MCTS not to be the first choice in video games is the degree of unpredictability introduced by stochastic factors. The game studios often require all interactions to be thoroughly tested before a game is released. Developing more formal methods of predicting and controlling the way MCTS-driven bots will behave in new scenarios in the game seems to be another promising future direction. 

The third impediment of MCTS application to video games is the difficulty of  implementing a forward model able to simulate games in a combinatorial-like fashion \citep{preuss2020games}. This issue has already been addressed by using simplified (proxy) game models and letting the MCTS algorithm operate either on abstraction (simplification) of the actual game or on some subset of the game (using a relaxed model).

In the view of the above, we believe that some methodological and/or computational breakthroughs may be required before MCTS  becomes a state-of-the-art (or sufficiently efficient) approach in RTS games domain.

In summary, the main advantages of MCTS which differentiate this method from the majority of other search algorithms can be summarized as follows. 
First of all, the problem tree is built asymmetrically, with the main focus on the most promising lines of action and exploring options that may most likely become promising. Second of all, the UCT formula is the most popular way of assuring the \emph{exploration vs. exploitation} balance. In practice, the method has proven efficient in various (though not all) search problems. Its additional asset is the theoretical justification of convergence.
Furthermore, the method is inherently easy to parallelize (cf. Section~\ref{sec:parallelization}). And finally, MCTS is the so-called \emph{anytime} algorithm, i.e. it can be stopped at any moment and still return a reasonably good solution (the best one found so-far). 
 
On the downside, since MCTS is a tree-search algorithm, it shares some weaknesses with other space-search. For instance, it is prone to search space explosion.
MCTS modifications, such as model simplifications or reducing the actions space are one way to tackle this problem. A different idea is to use MCTS as an \emph{offline} teacher that can be used  for sufficiently long time and generate high-quality training data.  Yet another method that is not search-based but rather learning-based and inherently capable of generalizing knowledge (e.g. an ML model) is trained on these MCTS-generated training samples. Consequently, a computationally heavy process is run just once (\emph{offline}) and then this time-efficient problem representation can be used in subsequent \emph{online} applications. The approach of combining an MCTS trainer with a fast learning-based representation can be hybridised in various ways, specific to a particular problem / domain of interest~\citep{kartal2019action,guo2014deep,soemers2019learning}.

Finally, we would like to stress the fact that MCTS, originally created for \emph{Go}, was for a long time devoted to combinatorial games and only recently started to ``expand'' its usage beyond this domain. We hope that the researchers working with other genres of games or in other fields will make more frequent attempts at  MCTS utilization in their domains, perhaps inspired by the MCTS modifications discussed in this survey.

\bibliography{mcts_2012-2020}{}
\bibliographystyle{spbasic}
\end{document}